\documentclass[twoside,11pt]{article}

\usepackage{jmlr2e}
\RequirePackage{amsmath}
\RequirePackage{fix-cm}
\usepackage[utf8]{inputenc} 
\usepackage[T1]{fontenc}    
\usepackage{hyperref}       
\usepackage{url}            
\usepackage{booktabs}       
\usepackage{amsfonts,amssymb}       
\usepackage{nicefrac}       
\usepackage{microtype}      
\usepackage{multirow}
\usepackage{xcolor}
\usepackage{makecell}
\usepackage{newtxtext,newtxmath}
\usepackage{bbm}
\usepackage{diagbox}
\usepackage{graphicx}
\usepackage{natbib}
\usepackage{algorithm}
\usepackage[algo2e]{algorithm2e}
\usepackage{algorithmic}
\usepackage{marvosym}
\usepackage{bbm}
\usepackage{epstopdf}
\usepackage{amssymb}
\usepackage{color}
\usepackage{colortbl}
\usepackage{natbib}
\usepackage{array}
\usepackage{extarrows}
\usepackage{ulem}
\newtheorem{Definition}{Definition}
\newtheorem{Proposition}{Proposition}
\newtheorem{Theorem}{Theorem}
\newtheorem{Lemma}{Lemma}

\newtheorem{Remark}{Remark}

\def\app#1#2{%
  \mathrel{%
    \setbox0=\hbox{$#1\sim$}%
    \setbox2=\hbox{%
      \rlap{\hbox{$#1\propto$}}%
      \lower1.1\ht0\box0%
    }%
    \raise0.25\ht2\box2%
  }%
}
\def\approxprop{\mathpalette\app\relax}
\usepackage{tikz}
\newcommand*{\circled}[1]{\lower.7ex\hbox{\tikz\draw (0pt, 0pt)%
    circle (.5em) node {\makebox[0em][c]{\small #1}};}}
\usepackage{etoolbox}
\usepackage{pifont}

\jmlrheading{1}{2020}{1-48}{4/00}{10/00}{}{}

\firstpageno{1}
\usepackage{lastpage}
\jmlrheading{22}{2021}{1-\pageref{LastPage}}{4/20; Revised
12/21}{12/21}{20-315}{Yuangang Pan, Ivor W. Tsang, Weijie Chen, Gang Niu and Masashi Sugiyama}
\ShortHeadings{Fast and Robust Rank Aggregation against 
	Model Misspecification}{Pan, Tsang, Chen, Niu and Sugiyama}
\begin{document}

\title{Fast and Robust Rank Aggregation against\\ 
	Model Misspecification}

\author{\name Yuangang Pan\thanks{Preliminary work was done during an internship at RIKEN AIP.} \email Yuangang.Pan@gmail.com \\
       \addr Center for Frontier AI Research\\  
       Agency for Science, Technology and Research (A*STAR)\\
       Singapore\\
       and\\
       \addr Australian Artificial Intelligence Institute\\
       University of Technology Sydney\\
       NSW 2007, Australia
       \AND
       \name Ivor W. Tsang \email Ivor.Tsang@gmail.com \\
       \addr Center for Frontier AI Research\\  
       Agency for Science, Technology and Research (A*STAR)\\
       Singapore\\
       and\\
       \addr Australian Artificial Intelligence Institute\\
       University of Technology Sydney\\
       NSW 2007, Australia
       \AND
       \name Weijie Chen \email wjcper2008@126.com \\
       \addr Zhijiang College\\
       Zhejiang University of Technology\\
       Hangzhou 310014, Zhejiang, China
       \AND
       \name Gang Niu \email gang.niu@riken.jp \\
       \addr Center for Advanced Intelligence Project\\
       RIKEN, Tokyo, 103-0027, Japan
       \AND
       \name Masashi Sugiyama \email sugi@k.u-tokyo.ac.jp \\
       \addr Center for Advanced Intelligence Project\\
       RIKEN, Tokyo, 103-0027, Japan \\
       and\\
       \addr Graduate School of Frontier Sciences\\
       University of Tokyo\\
       Chiba 277-8561, Japan
       }
\editor{Sathiya Keerthi}

\maketitle

\begin{abstract}
In rank aggregation (RA), a collection of preferences from different users are summarized into a total order under the assumption of homogeneity of users. Model misspecification in RA arises since the homogeneity assumption fails to be satisfied in the complex real-world situation. Existing robust RAs usually resort to an augmentation of the ranking model to account for additional noises, where the collected preferences can be treated as a noisy perturbation of idealized preferences. Since the majority of robust RAs rely on certain perturbation assumptions,  they cannot generalize well to agnostic noise-corrupted preferences in the real world. In this paper, we propose CoarsenRank, which possesses robustness against model misspecification. Specifically, the properties of our CoarsenRank are summarized as follows: 
(1) CoarsenRank is designed for mild model misspecification, which assumes there exist the ideal preferences (consistent with model assumption) that locates in a neighborhood of the actual preferences. 
(2) CoarsenRank then performs regular RAs over a neighborhood of the preferences instead of the original dataset directly. Therefore, CoarsenRank enjoys robustness against model misspecification within a neighborhood. 
(3) The neighborhood of the dataset is defined via their empirical data distributions. Further, we put an exponential prior on the unknown size of the neighborhood, and derive a much-simplified posterior formula for CoarsenRank under particular divergence measures.
(4) CoarsenRank is further instantiated to Coarsened Thurstone, Coarsened Bradly-Terry, and Coarsened Plackett-Luce with three popular probability ranking models. Meanwhile,  tractable optimization strategies are introduced with regards to each instantiation respectively. 
In the end, we apply CoarsenRank on four real-world datasets. Experiments show that CoarsenRank is fast and robust, achieving consistent improvements over baseline methods.
\end{abstract}

\begin{keywords}
Robust Rank Aggregation, Model Misspecification, CoarsenRank, Coarsened Bradly-Terry, Coarsened Plackett-Luce
\end{keywords}

\section{Introduction}\label{Intro}
Rank aggregation (RA) refers to the task of recovering the total order over a set of items, given a collection of pairwise/partial/full preferences over items~\citep{lin2010rank}. Therefore, RA is a practical and useful approach to summarize user preferences~\citep{de1781memoire}. 
Preferences could arise not only by explicitly querying users but also through passive data collection, i.e., by observing user purchasing behavior~\citep{baltrunas2010group}, clicks on search engine results~\citep{dwork2001rank}, etc. Compared to rating items, the preferences are more natural expressions of user opinions which can provide more consistent results~\citep{raman2014methods}. The flexible collection of preferences enables successful application of rank aggregation in various fields, from image rating~\citep{liang2014beyond} to document recommendation~\citep{sellamanickam2011pairwise}, peer grading~\citep{raman2014methods}, opinion analysis~\citep{chatterjee2018weighted}, bioinformatics \citep{kim2014hydra} and mental fatigue monitoring~\citep{PANNC2020,PANNC2021}. 

A basic assumption underlying the vanilla RA is that all preferences are provided by homogeneous users, sharing the same annotation accuracy and agreeing with the single ground truth ranking~\citep{dwork2001rank,li2017comparative,chiang2017rank, li2019online}. However, this homogeneity assumption is rarely satisfied due to the flexible data construction and the complex real-world situation \citep{gormley2005exploring,kolde2012robust,mollica2017bayesian,li2018learning}. For example, the reliability of each user may not be necessarily the same due to the diverse background of each user regarding the candidate items. Moreover, the users are more likely coming from a heterogeneous community and the single total order assumption is no longer suitable since different users judge the items from different perspectives. Therefore, RA usually suffers from model misspecification, namely the inconsistency between the collected ranking data and the homogeneity assumption of RA~\citep{Pan2018rank}. 

To address the above inconsistency issue, existing robust RAs resort to an augmentation of the ranking model to account for additional perturbation, where the collected preferences are viewed as a noisy perturbation of some idealized preferences~\citep{han2018robust}.  Particularly, \cite{chen2013pairwise} studied RA in a crowdsourcing environment and proposed CrowdBT for noisy pairwise preferences. CrowdBT models the user reliability with one extra parameter, following the two-coin Dawid-Skene model~\citep{raykar2010learning}. \cite{han2018robust} proposed ROPAL, which extended CrowdBT for noisy partial preference using more parameters. Note that ROPAL requires a few ground truth preferences from each user for initializing the parameters. It constraints their method to a crowdsourcing setting,  where multiple preferences from each user are available.  \cite{raman2014methods} introduced a general framework, called PeerGrader, to aggregate ordinal peer gradings from peer graders while exploring each grader's reliability by introducing a scale factor. However, each user usually provides one preference in real applications, which would cause overfitting since it needs to estimate the reliability w.r.t. each preference~\citep{sajjadi2016peer}. The same problem also arises in \cite{xu2017exploring}, which formulates the robust RA as outlier detection and introduces a deviation factor for each preference to account for the unknown noise-perturbation. Indeed, these previous attempts simply amount to convolving the original ranking model with some pre-assumed perturbation mechanism. It leads to a new model with a few more parameters but is just as bound to be misspecified w.r.t. other overlooked perturbations.

The above analysis motivates us to present a novel robust RA approach, called CoarsenRank. The main idea of CoarsenRank is to perform regular RA over a neighborhood of the collected preferences, which enables CoarsenRank against mild model misspecification within the defined neighborhood~\citep{volpi2018generalizing,chen2018robust}.
However, it is usually intractable to infer directly over the neighborhood of the ranking data because of the unlimited samples involved. Further, it also prohibits sampling-based stochastic gradient solutions in the optimization community due to the particularity of the ranking data. Inspired by~\cite{miller2018robust}, which avoids inferring over the neighborhood of the dataset by transforming the problem into a tractable fractional likelihood formulation~\citep{bhattacharya2019bayesian}. For the sake of tractability, the neighborhood of the dataset is first defined as the neighborhood of its empirical data distribution. In particular, the relative entropy is adopted as the divergence metric due to its simplicity~\citep{ben2013robust,namkoong2017variance}. 
We further introduce a prior distribution for the unknown size of the neighborhood to avoid parameter tuning and derive a much-simplified formula for CoarsenRank.

More precisely, we summarize our main contributions in the following:
\begin{itemize}
	\item We introduce a novel robust rank aggregation method called CoarsenRank. CoarsenRank performs RA over the neighborhood of the ranking data instead of original dataset directly. To our best knowledge, CoarsenRank is the first rank aggregation method against model misspecification and enjoys distributional robustness.
	\item We obtain a computationally efficient formula for CoarsenRank, which introduces only one extra hyperparameter to vanilla ranking models.  Further, we instantiate CoarsenRank with three popular probability ranking models and analyze the optimization strategies, respectively. 
	To avoid hyperparameter tuning, an efficient model selection method is introduced to choose the single hyperparameter in a data-driven manner.
	\item We successfully applied our CoarsenRank on four real-world datasets. Empirical results demonstrate (1) CoarsenRank shows superior reliability against agnostic noises over existing robust RA methods, especially when the number of annotations per user is insufficient; (2) CoarsenRank enjoys linear algorithm complexity, which has great potential for a large-scale scenario.
\end{itemize}
The rest of this paper is organized as follows. Section~\ref{Sect_2} discusses RA under model misspecification and introduces the main idea of our CoarsenRank. 
In Section~\ref{Sect_3}, we pave the theoretical foundation for CoarsenRank and illustrate how CoarsenRank enables us to perform robust RA against model misspecification. Section~\ref{Sect_4} presents an efficient EM algorithm as well as a Gibbs sampling algorithm for CoarsenRank and discusses a data-driven strategy for hyperparameter selection. Section~\ref{related-work} summarizes the differences between our CoarsenRank and related (robust) RA models.  Section~\ref{experiment} demonstrates the efficacy of CoarsenRank through empirical results on four real-world datasets. Section~\ref{Conclusion} concludes the paper and envisions future work. 

\section{Problem statement and literature review}\label{Sect_2}

In this section, we first introduce the problem setting of vanilla RA and  RA under model misspecification. Furthermore, we summarize previous robust RA for alleviating the model misspecification, as well as their deficiencies.  Then, we motivate our Coarsened RA, which perform regular RA over a neighborhood of the collected preferences, and therefore enjoys distributional robustness against noise-agnostic perturbation within a neighborhood. 

\subsection{Rank aggregation}
In Table~\ref{Math_notation}, we first illustrate the common mathematical notations that are used later. 

\begin{table}[!t]
\large
\centering
\caption{\label{Math_notation} Common mathematical notations}
\renewcommand{\arraystretch}{1.3}
\setlength{\tabcolsep}{1.3mm}{	
\scalebox{0.82}{
\begin{tabular}{l|l}
\toprule[1.1pt]
Notation         & Explanation                           \\\hline
$M$         & number of items \\
$N$         & number of preferences\\
$k$         & length of preferences, which could be variant with regards to each preference\\
$\mathcal{O}$    & set of items, $\mathcal{O} = \{o_1, o_2, \ldots, o_M\}$  \\
$o_i > o_j$  &  item~$o_i$ is preferred over item~$o_j$\\
$\rho_n$         & $n$-th real preference, $\rho_n: \rho_n^1> \rho_n^2> \ldots> \rho_n^k$, $\{\rho_n^1, \rho_n^2, \ldots, \rho_n^k\} \subseteq  \mathcal{O}$   \\
$\varrho_n$  & $n$-th idealized preference, $\varrho_n: \varrho_n^1> \varrho_n^2> \ldots> \varrho_n^k$, $\{\varrho_n^1, \varrho_n^2, \ldots, \varrho_n^k\} \subseteq  \mathcal{O}$ \\
$\mathcal{R}_N$  & collection of real preferences, $\mathcal{R}_N = \{\rho_1, \rho_2, \ldots, \rho_N\}$  \\
$\Re_N$          &collection of idealized preferences,  $\Re_N= \{\varrho_1, \varrho_2, \ldots,\varrho_N\}$, satisfying the homogeneity assumption   \\
$B_s(\mathcal{R}_N,  \epsilon)$ & sample-level neighborhood of the ranking dataset $\mathcal{R}_N$\\
$B_d(\mathcal{R}_N,  \epsilon)$ & distribution-level neighborhood of the ranking dataset $\mathcal{R}_N$\\
$P_\theta$   &  probability ranking model, $\theta$ denotes the model parameter \\
$P_o$        &  real preference generation distribution  \\
$F_N(x|\mathcal{R}_N)$ &  the empirical distributions of the real ranking dataset $\mathcal{R}_N$ \\
$F_N(x|\Re_N)$ & the empirical distributions of the idealized ranking dataset $\Re_N$\\
$\Theta$  &  space of possible parameter values that defines a ranking model $P_\theta$\\
$\mathcal{P}$ & set of all probability rank models with parameterization space $\Theta$, $P_\theta \in \mathcal{P}$ and $\theta \in \Theta$ \\
$D(\cdot, \cdot)$ & divergence metric between two datasets, defined via their empirical data distributions\\
$\mathbb{I}_x(y)$ & indicator function, which is one at $x = y$ or zero otherwise\\ \toprule[1.1pt]
\end{tabular}}}
\end{table}

\begin{figure}[!htb]
	\centering
	\includegraphics[width=0.85\textwidth]{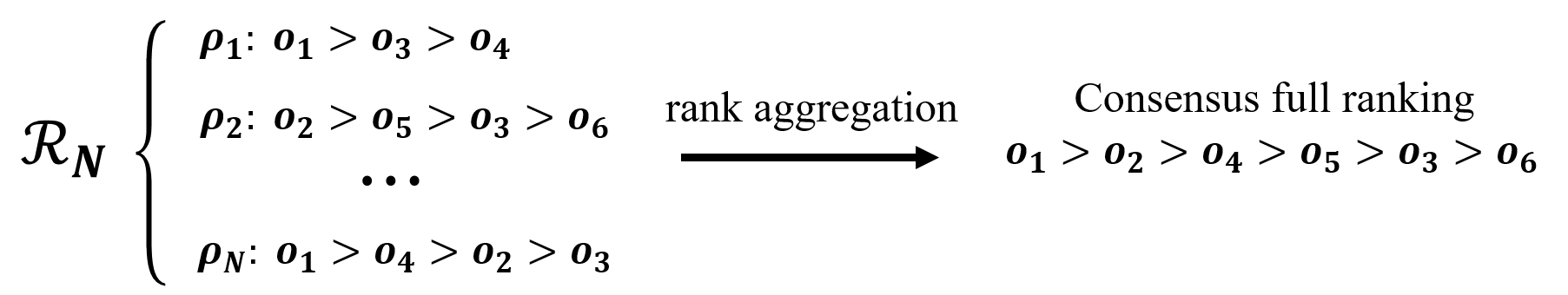}
	\caption{\label{Vanilla_RA} The paradigm of rank aggregation.}
\end{figure}

Let $\mathcal{R}_N$ denote a collection of partial preferences $\{\rho_1, \rho_2, \ldots, \rho_N\}$ over the item set~$\mathcal{O} = \{o_1, o_2, \ldots, o_M\}$. 
The goal of rank aggregation is then to aggregate the collected preferences~$\mathcal{R}_N$ into a consensus order over all $M$ items in $\mathcal{O}$ (See Figure~\ref{Vanilla_RA}). The consensus order should achieve the maximum agreement among all preferences in $\mathcal{R}_N$~\citep{dwork2001rank}. 

In this paper, we focus our work on rank aggregation using the probability ranking model. Particularly, it assumes there exists a preference generative model~$P_o$ from which the preferences~$\mathcal{R}_N$ are sampled, i.e.,  $\mathcal{R}_N = \{\rho_n | \rho_n \sim P_o, n = 1, 2, \ldots, N\}$. However, the real data generation model $P_o$ is hardly accessible due to the complexity of the real situation. For the sake of easier modeling, a parameterized rank model $P_{\theta}$ is usually adopted under the assumption of homogeneity of users\footnote{Sampling the partial preferences from a specific probability ranking model is not our focus in this paper. Please refer to~\cite{liu2019learning,zhao2019learning} for related literature.}. Let $\mathcal{P}$ be the set of all probability rank models under the homogeneity assumption. For the sake of easier modeling, a parameterized rank model $P_{\theta}$ is usually adopted under the assumption of homogeneity of users. Then a maximum likelihood estimation (MLE) for RA can be formulated as follows,

\begin{equation}\label{RA_vanilla}
\max_{\theta \in \Theta}\ P_\theta (\mathcal{R}_N), \quad \mathrm{where} \quad P_\theta \in \mathcal{P} \ \text{and} \ \mathcal{R}_N = \{\rho_n | \rho_n \sim P_o, n = 1, 2, \ldots, N\}.
\end{equation}
$P_\theta (\mathcal{R}_N) = \prod_{n =1}^N P_\theta(\rho_n)$ denotes the likelihood over the collected preferences $\mathcal{R}_N$.  $P_\theta$ is usually instantiated with Thurstone model~\citep{thurstone1927law,thurstone1927method}, Bradley-Terry model~\citep{bradley1952rank},  Plackett-Luce model~\citep{plackett1975analysis,luce1959individual},~etc.  Note that the model parameter $\theta$ is usually associated with each item, where the full ranking list could be derived accordingly after $\theta$ is inferred. For example, the full ranking list can be obtained by sorting the model parameter $\theta$ in the case of the Thurstone/Bradley-Terry/Plackett-Luce model.

\subsection{Rank aggregation under model misspecification}

In this section, we discuss RA under model misspecification. The term ``model misspecification'' here refers to the mismatch between the ranking model $P_\theta$ and the ranking dataset $\mathcal{R}_N$, namely the collected user preferences do not strictly satisfy the user homogeneity assumption of the ranking model.

\begin{figure}[!tb]
	\centering
	\includegraphics[width=1\textwidth]{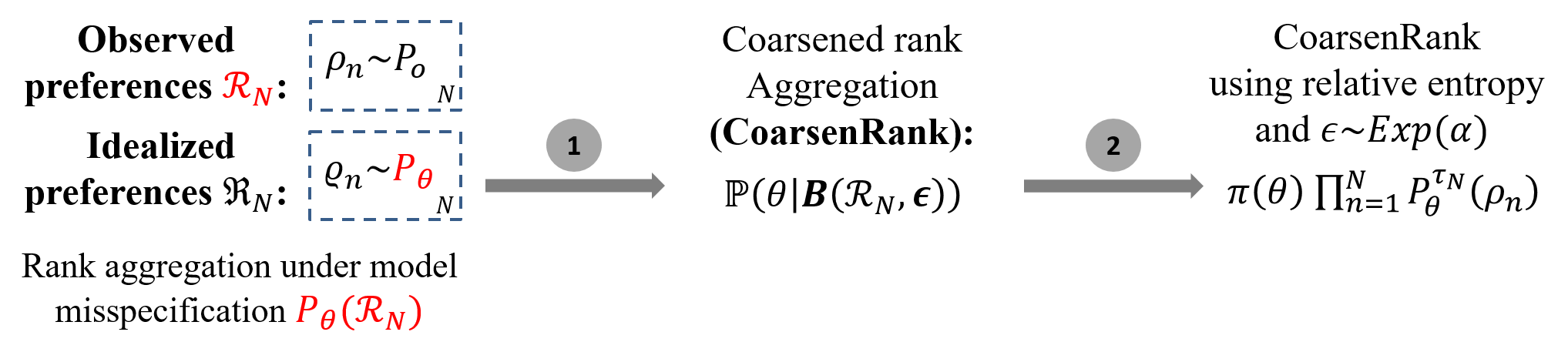}\vskip-0.1in
	\caption{\label{Coarsen}The logic stream of our CoarsenRank. Stage~1: performing rank aggregation over a neighborhood of the collected preferences (See~Equation~\eqref{Coarsen_RA} and Equation~\eqref{Coarsen_RA_Compact}). Stage~2: adopting relative entropy as the divergence measure and assign an exponential prior for the size of the neighborhood (See Theorem~\ref{extension_1} and Theorem~\ref{extension_2}).}\vskip-0.1in
\end{figure}

The model misspecification would arise when preferences were not strictly collected from a homogeneous user community due to the flexible data construction and the complex real situation (See Figure~\ref{Coarsen}). For example, the reliability of each user would not be the same and the single total order assumption would be no longer satisfied. Mathematically, we adopt the parameterized ranking model $P_\theta \in \mathcal{P}$ under the homogeneity assumption, while the real preference generation distribution $P_o$ violates this assumption, i.e., $P_o \notin \mathcal{P}$. Therefore, an MLE for RA under model misspecification can be formulated as follows,

\begin{equation}\label{RA_mis}
\begin{aligned}
\max_{\theta \in \Theta} \quad P_\theta (\mathcal{R}_N)
\quad \mathrm{where} \quad P_\theta \in \mathcal{P}, \ P_o \notin \mathcal{P} \ \text{and} \  \mathcal{R}_N = \{\rho_n | \rho_n \sim P_o, n = 1, 2, \ldots, N\}.
\end{aligned}
\end{equation}
For the sake of explanation, let $\Re_N$ represent a virtual dataset $\{ \varrho_n | \varrho_n \sim P_{\theta}, n = 1, 2, \ldots, N\}$, which consists of idealized preferences and satisfies the homogeneity assumption. Then, RA under model misspecification can be formulated as noisy RA, where the collected preferences are viewed as a noisy perturbation of some idealized preferences. 

Then, we come to robust rank aggregation against model misspecification, namely how to achieve a reliable total order from the collected preferences $\mathcal{R}_N$ using a misspecified ranking model $P_\theta$. 

\subsection{Previous attempts: convolving ranking model with certain perturbation mechanisms}

When encountering model misspecification,  a remedy solution for accessing a correct rank model could be
\begin{equation}\label{RA_Correction}
P_{\theta}(\rho_n) =  \sum_{\varrho_n}
P_{\theta}(\varrho_n) P(\rho_n|\varrho_n ) =  \sum_{\varrho_n} P(\rho_n, \varrho_n), \quad \text{where} \ \rho_n \sim P_o, \ \varrho_n \sim P_{\theta}.  
\end{equation}
Considering the discrete characteristics of ranking space, the ranking distribution i.e., $P_o$ and $P_{\theta}$ as well as the conditional distribution $P(\rho_n|\varrho_n )$ and joint distribution $P(\rho_n, \varrho_n )$  should all be the discrete distribution.

Previous approaches usually resort to an augmentation of the ranking model to account for additional error/noise/uncertainty caused misspecification. For the sake of tractability, the perturbation mechanism is usually defined at the sample level. According to Equation~\eqref{RA_Correction},  there are essentially two ways of implementing this:
\begin{itemize}
    \item One intuitive approach is to correct each preference by pre-assuming some perturbation distribution, i.e., $P(\rho_n|\varrho_n )$. However, this simply amounts to convolving the original model distribution $P_\theta$ with the predefined perturbation, leading to a new model that has a few more parameters but is just as bound to be misspecified w.r.t. other overlooked perturbations. 
    
    \item The second approach would be to model the joint distribution $P(\rho_n, \varrho_n)$ directly, which needs to take into consideration all potential perturbations. Essentially, it needs to be a nonparametric model for $P(\rho_n, \varrho_n)$, but would easily be computationally intractable.
\end{itemize}

Meanwhile, the perturbation patterns leading to model misspecification vary from setting to setting. It is impossible to design a universal practice that can be generalized to most settings. Therefore, in this paper,  we perform rank aggregation against model misspecification from another perspective. 

\subsection{Our CoarsenRank: rank aggregation over the neighborhood of ranking data}

In many situations, it is impractical to correct the model, and these are the situations our method is intended to address. We are concerned with robust rank aggregation against model misspecification (Equation~\eqref{RA_mis}) in general, not just one particular kind of perturbation considered in previous work.  Recent advances of robust Bayesian inference~\citep{miller2018robust,volpi2018generalizing} raised the Coarsening mechanism, namely \textit{inferring over the neighborhood of the original dataset would equip the learning model with distributional robustness}. Motivated by the proposed Coarsening mechanism, we consider performing rank aggregation over the neighborhood of the ranking data.

To deliver our model, we first give the definition of the neighborhood in the sense of ranking data as follows, 
\begin{Definition}[sample-level neighborhood]
Let $\mathcal{R}_N$ denote the ranking dataset and $\rho$ represent one preference belonging to $\mathcal{R}_N$. We define the neighborhood $B_s(\mathcal{R}_N,  \epsilon)$ of the $\mathcal{R}_N$ with size $\epsilon > 0$ as follows: 
\begin{equation}\label{sample_level}
B_s(\mathcal{R}_N,  \epsilon) = \{ \rho' | D(\rho', \rho) < \epsilon, \ \exists \rho \in \mathcal{R}_N\},
\end{equation}
where $D(\cdot, \cdot)$ denotes some distance measure between two preferences $\rho'$ and $\rho$, e.g., Kendall tau distance~\citep{kendall1938new}, Spearman's rank correlation~\citep{daniel1990applied}.
\end{Definition}

\begin{Definition}[distribution-level neighborhood]
Let $\mathcal{R}_N$ denote the ranking dataset and $B_d(\mathcal{R}_N,  \epsilon)$ denote the neighborhood of the ranking dataset $\mathcal{R}_N$ with size $\epsilon > 0$. Then, we define
\begin{equation}\label{distribution_level}
B_d(\mathcal{R}_N,  \epsilon) = \{ \mathcal{R}'_N | D(\mathcal{R}'_N, \mathcal{R}_N) < \epsilon\},
\end{equation}
where $D(\cdot, \cdot)$ denotes some distance measures between two ranking datasets. The distance measure between two datasets is usually defined as the divergence of their corresponding empirical distributions. Popular divergence measures between distributions are Kullback-Leibler (KL) divergence~\citep{kullback1951information}, $f$-divergence~\citep{ali1966general} and Wasserstein metric~\citep{villani2008optimal}.
\end{Definition}

\begin{Proposition} Given any ranking dataset $\mathcal{R}_N$, $\mathcal{R}_N$ must be (1) a subset of its sample-level neighborhood $B_s(\mathcal{R}_N,  \epsilon)$ if $D(\cdot, \cdot)$ is defined between two preferences, or (2) an element of its distribution-level neighborhood $B_d(\mathcal{R}_N,  \epsilon)$ if $D(\cdot, \cdot)$ is defined between two ranking datasets. Namely
\begin{equation*}
\begin{aligned}
\qquad \qquad \qquad \qquad \qquad \mathcal{R}_N  & \subseteq B_s(\mathcal{R}_N,  \epsilon),\qquad \qquad \qquad \textbf{sample-level neighborhood,}\\
\qquad \qquad \qquad \qquad \quad \mathcal{R}_N  & \in  B_d(\mathcal{R}_N,  \epsilon), \qquad \qquad \textbf{distribution-level neighborhood.}
\end{aligned}
\end{equation*}
\end{Proposition}
\paragraph{Proof:} 
In terms of sample-level neighborhood, we have 
\[
\forall \rho \in \mathcal{R}_N, \quad \exists \rho^{*} = \rho, \ \text{s.t.} \ D(\rho^{*}, \rho) = 0 < \epsilon. 
\]
Then $\rho \in B_s(\mathcal{R}_N,  \epsilon)$ holds. Accordingly,  $\mathcal{R}_N  \subseteq B_s(\mathcal{R}_N,  \epsilon)$ holds according to the definition of sample-level neighborhood in Equation~\eqref{sample_level}.

In terms of distribution-level neighborhood, we have 
\[
D(\mathcal{R}_N, \mathcal{R}_N) = 0 < \epsilon. 
\]
Therefore, we have $\mathcal{R}_N  \in B_d(\mathcal{R}_N,  \epsilon)$ hold according to the definition of distribution-level neighborhood in Equation~\eqref{distribution_level}. 

Note that the proof is valid for any particular choice of the distance metric,  either sample level or distribution level. \hfill\rule{2mm}{2mm}

\begin{Definition}[empirical data distribution] Let $F_N(x|\mathcal{R}_N)$ and $F_N(x|\Re_N)$ denote the empirical distributions of the ranking datasets $\mathcal{R}_N$ and $\Re_N$, respectively.
\begin{equation}\label{empirical_data_distribution}
\begin{aligned}
F_N(x|\mathcal{R}_N) & =  \frac{1}{N}  \sum_{n=1}^N  \mathbb{I}_{\rho_n}(x),\quad \mathrm{where} \ \mathbb{I}_{\rho_n}(x) = \left\{\begin{array}{ll}{1,} & {x=\rho_n} \\ {0,} & {x \neq \rho_n}\end{array}\right.,\\
F_N(x|\Re_N) & =  \frac{1}{N}  \sum_{n=1}^N  \mathbb{I}_{\varrho_n}(x),\quad \mathrm{where} \ \mathbb{I}_{\varrho_n}(x) = \left\{\begin{array}{ll}{1,} & {x=\varrho_n} \\ {0,} & {x \neq \varrho_n}\end{array}\right. .
\end{aligned}
\end{equation}
In this paper, we assume that the empirical distribution converges to the corresponding preference generation distribution, namely $F_N(x|\mathcal{R}_N) \rightarrow P_o$ and $F_N(x|\Re_N) \rightarrow P_\theta$ when $N \rightarrow +\infty$. 
\end{Definition}

For the sake of brevity, we introduce our work following the definition of distribution-level neighborhood. Let $\Re_N \sim P_\theta$ denote every preference of the ranking dataset $\Re_N $ is sampled from the ranking distribution $P_\theta$, namely $\forall \varrho_n \in \Re_N,  \varrho_n \sim P_{\theta}$. We assume that \textit{the idealized ranking dataset $\Re_N$ locates in the small neighborhood of the actually collected preferences~$\mathcal{R}_N$}, i.e., 
\[
 \exists \text{ a small } \epsilon, \ \forall \Re_N\sim P_\theta, \ \text{we have} \ \Re_N \in  B_d(\mathcal{R}_N,  \epsilon)
\]
This is a basic assumption in distributional robustness literature~\citep{chen2018robust}. Otherwise, if the size of neighborhood $\epsilon$ which satisfies our assumption is very large, it means a completely wrong model is adopted and it is impossible to learn a meaningful result. This is why we call our setting ``mild model misspecification''. Meanwhile, the sense of ``neighborhood'' in the distribution level covers most types of noise perturbations~\citep{chen2018outlier}. Therefore, the MLE of our Coarsened rank aggregation (CoarsenRank) can be formulated as follows,
\begin{equation}\label{Coarsen_RA}
\max_{\theta \in \Theta}
P_\theta (\Re_N), \quad \mathrm{where} \ \Re_N\sim P_\theta \ \text{and} \ \Re_N \in  B_d(\mathcal{R}_N,  \epsilon).
\end{equation}
Note that we use the word ``Coarsen'' to emphasize the learning paradigm which pursues the distributional robustness by inferring over the neighborhood of the dataset~\citep{miller2018robust}. 

An equivalent (but compact) formulation of our CoarsenRank (Equation~\eqref{Coarsen_RA}) can be derived as follows
\begin{equation}\label{Coarsen_RA_Compact}
\begin{aligned}
\text{Equation~\eqref{Coarsen_RA}} 
\overset{\circled{1}}{\Longleftrightarrow} &
\max_{\theta \in \Theta} \mathbb{E}_{\Re_N \sim P_\theta}\mathbb{P} (\Re_N|\theta, D(\mathcal{R}_N, \Re_N) < \epsilon)\\
\overset{\circled{2}}{\Longrightarrow} &
\max_{\theta \in \Theta} \mathbb{E}_{\Re_N \sim P_\theta}\mathbb{P}(\theta |  D(\mathcal{R}_N, \Re_N) < \epsilon),
\end{aligned}
\end{equation}
where $\circled{1}$ is an equivalent MLE formulation rearranged according to the definition of distribution-level neighborhood (Equation~\eqref{distribution_level}). By assigning a suitable prior for model parameter $\theta$, i.e., $\theta \sim \pi(\theta)$, $\circled{2}$ deduces the maximum a posteriori probability (MAP) estimate of our CoarsenRank model. 

Equation~\eqref{Coarsen_RA_Compact} reveals that: (1)~our CoarsenRank degenerates to vanilla rank aggregation method (Equation~\eqref{RA_vanilla}) when the collected preferences satisfy the homogeneity assumption; (2) our CoarsenRank would be robust to noise-agnostic perturbations as long as the idealized preferences locates in the neighborhood of the collected preferences when model misspecification arises; and (3) our CoarsenRank would fail to output a reliable total ranking list when the collected preferences significantly violate the homogeneity assumption. The same is true for other vanilla rank aggregation methods and most of robust RAs which fail to capture this perturbation. Therefore, compared with the previous methods, our CoarsenRank is robust to most potential perturbations within a neighborhood, not only to some pre-assumed perturbations. 

\begin{Remark}[Coarsening mechanism VS. Minimax distributional robustness]\label{Coarsen_mechanism}
The Coarsening mechanism $\max_{\theta \in \Theta}
\mathbb{E}_{\Re_N \in B_d(\mathcal{R}_N,  \epsilon)}P_\theta (\Re_N)$ shares a similar  formula with the minimax distributional robustness $\min_{\theta \in \Theta}
\max_{\Re_N \in B_d(\mathcal{R}_N,  \epsilon)}\ell(\Re_N|\theta)$~\citep{sinha2017certifying}. $\ell(\theta)$ denotes the loss function, which is usually replaced with the negative log-likelihood. The Coarsening mechanism  aims to maximize the likelihood over the neighborhood of original dataset $\mathcal{R}_N$. The minimax distributional robustness aims to minimize the loss using the worst data samples in the neighborhood of original dataset $\mathcal{R}_N$. The neighborhood in the Coarsening mechanism is usually defined at the distribution-level (Equation~\eqref{distribution_level}) where a Bayesian criterion can be adopted to estimate the proper size of the neighborhood for each dataset; while the neighborhood in the minimax distributional robustness is usually defined at the sample-level (Equation~\eqref{distribution_level}) and a fixed size neighborhood is adopted once for all. The choice of distance measures $D(\cdot, \cdot)$ influences robustness guarantee and tractability in the both two paradigms.\hfill\rule{2mm}{2mm}
\end{Remark}

\begin{Remark}[Coarsening mechanism VS. Rank-dependent coarsening]
The word Coarsening also rises in~\cite{fahandar2017statistical}, which, however, has a totally different meaning there. The term ``rank-dependent coarsening'' refers to the process of turning a full ranking into an incomplete one~\citep{fahandar2017statistical}. It is different from our ``Coarsening mechanism'', which refers to the paradigm that performing the Bayesian inference over the neighborhood of the original dataset.
\hfill\rule{2mm}{2mm}
\end{Remark}

\section{Coarsened rank aggregation}\label{Sect_3}

In this section, we first illustrate how CoarsenRank enables us to perform robust rank aggregation against model misspecification. Meanwhile, a simplified formula is derived for CoarsenRank, which introduces only one extra hyperparameter to vanilla ranking models. Then, we instantiate our CoarsenRank framework with three popular probability ranking models and analyze their optimization strategies, respectively.

\subsection{Distributional robustness of Coarsening mechanism}

Assuming that the empirical distribution defined in Equation~\eqref{empirical_data_distribution} converges to the corresponding data generating distribution, namely $F_N(x|\mathcal{R}_N) \rightarrow P_o$ and $F_N(x|\Re_N) \rightarrow P_\theta$ when $N \rightarrow +\infty$, we come to Theorem~\ref{fundation_theory}. Note that this result is essentially S3.1 in the Supplement of~\cite{miller2018robust}. We also provide a proof in the Appendix for the sake of completeness.

\begin{Theorem}\label{fundation_theory}  Suppose $D(\mathcal{R}_N, \Re_N)$ is an almost surely-consistent estimator\footnote{In probability theory, an event happens almost surely if it happens with probability one.} of $D(P_o, P_\theta)$, namely $D(\mathcal{R}_N, \Re_N) \xlongrightarrow[N\rightarrow  +\infty]{\text{a.s.}} D(P_o, P_\theta)$, where $F_N(x|\mathcal{R}_N) \rightarrow P_o$ and $F_N(x|\Re_N) \rightarrow P_\theta$ when $N \rightarrow +\infty$. Assume $\mathbb{P}(D(P_o, P_\theta) = \epsilon) = 0$ and  $\mathbb{P}(D(P_o, P_\theta) < \epsilon) > 0$, then we have 
\begin{equation}\label{theory_1}
\mathbb{P}(\theta | D(\mathcal{R}_N, \Re_N) < \epsilon) \ \xlongrightarrow[N\rightarrow  +\infty]{\text{a.s.}}\ \mathbb{P}(\theta | D(P_o, P_\theta) < \epsilon),
\end{equation}
for any $\theta \in  \Theta$ such that $\int  |\theta| \mathbb{P}(d\theta) < \infty$.
\end{Theorem}

Theorem~\ref{fundation_theory} is a general conclusion in robust Bayesian inference \citep{miller2018robust}. It justifies our motivation to pursue robustness in a distributional sense. In what follows, we extend Theorem~\ref{fundation_theory} to some variants which possess nice properties for robust rank aggregation.

\subsubsection{Level of distributional robustness}
The value of the parameter $\epsilon$ denotes the level of deviation about the actually collected preferences from the idealized preferences, which varies from dataset to dataset. Simply fixing the $\epsilon$ to a small value, the Coarsening mechanism degenerates to a minimum-expectation problem as we discussed in Remark~\ref{Coarsen_mechanism}. It would be heuristic without sufficient prior knowledge about idealized preferences, since a small $\epsilon$ may fail to account for the unknown distribution deviation while a large $\epsilon$ means an exponential level of ranking space to search.  To ease the burden of pre-defining $\epsilon$, we treat it as a random variable and introduce a prior on it, where an efficient model selection method is introduced. In particular, we have the following conclusion. 

\begin{Theorem}\label{extension_1} Assume $\theta \sim \pi(\theta)$, the approximate posterior can be further simplified.
\begin{equation}\label{extend_to_exp}
\mathbb{P}(\theta | D(P_o, P_\theta) < \epsilon) = \frac{\pi(\theta) \mathbb{P}(D(P_o, P_\theta) < \epsilon | \theta)}{\int_{\theta} \pi(\theta)\mathbb{P}(D(P_o, P_\theta) < \epsilon | \theta)d\theta}  \propto   \mathrm{exp}(-\alpha D(P_o, P_\theta))\pi(\theta),
\end{equation}
when random variable $\epsilon$ subjects to an exponential prior, i.e., $\epsilon \sim \mathrm{Exp}(\alpha)$.
\end{Theorem}

\paragraph{Proof:} Note that since $\epsilon \sim \mathrm{Exp}(\alpha)$, we have 
\begin{equation*}
\begin{aligned}
\mathbb{P}(D(P_o, P_\theta) < \epsilon |\theta) & = 1 - \mathbb{P}(\epsilon \leq D(P_o, P_\theta)|\theta) \\
& = 1 - (1 - \mathrm{exp}(-\alpha D(P_o, P_\theta))) \\
& = \mathrm{exp}(-\alpha D(P_o, P_\theta)),
\end{aligned}
\end{equation*}
where the second equation holds because the cumulative distribution function $\mathbb{P}(D(P_o, P_\theta) > \epsilon |\theta)$ is independent of $\theta$~\citep{bishop2006pattern}. Then, we can substitute $\mathbb{P}(D(P_o, P_\theta) < \epsilon |\theta)$ in Equation~\eqref{extend_to_exp} with $\mathrm{exp}(-\alpha D(P_o, P_\theta))$ and complete the proof while omitting the normalization constant.~\hfill\rule{2mm}{2mm}

Indeed, a very large class of distributions can be adopted as the prior for $\epsilon$. A case of particular~interest arises when $\epsilon \sim\text{Exp}(\alpha)$, since it leads to a computationally simple formula via maintaining an exponential formulation. The efficacy of the exponential prior is verified in our experiment (See Section~\ref{experiment}). 

Inspired by the exponential formulation of the posterior derived in Equation~\eqref{extend_to_exp}, we give the following derivations (Equation~\eqref{Standard_Posterior}) to explain why the vanilla rank aggregation is lack of robustness.
\begin{align}\label{Standard_Posterior}
\mathbb{E}_{\Re_N \in B_d(\mathcal{R}_N,  0)}\mathbb{P}(\theta |\Re_N)&=\mathbb{P}(\theta | \mathcal{R}_N)   = \frac{\pi(\theta)P_\theta(\mathcal{R}_N)}{ \int \pi(\theta)P_\theta(\mathcal{R}_N) d\theta}  \propto \pi(\theta)P_\theta(\mathcal{R}_N)\\
& \overset{\circled{1}}{=} \pi(\theta)\mathrm{exp}(\sum_{n=1}^N \mathrm{log}P_\theta(\rho_n))\overset{\circled{2}}{=} \pi(\theta)\mathrm{exp}(N \sum_{n=1}^N F_N(\rho_n|\mathcal{R}_N) \mathrm{log} P_\theta(\rho_n))\nonumber\\
&  \overset{\circled{3}}{\approx} \pi(\theta)\mathrm{exp}(N \sum P_o \mathrm{log} P_\theta)\overset{\circled{4}}{ \propto } \pi(\theta)\mathrm{exp}(-N\mathcal{D}_{\mathrm{KL}}(P_o \| P_ \theta)),\nonumber
\end{align}
where $\circled{1}$ holds because $P_\theta (\mathcal{R}_N) = \prod_{n =1}^N P_\theta(\rho_n)$ is the likelihood. $\circled{2}$ holds following the definition of the empirical data distribution $F_N(x|\mathcal{R}_N) =  \frac{1}{N}  \sum_{n=1}^N  \mathbb{I}_{\rho_n}(x)$. $\circled{3}$~indicates Monte Carlo approximation. $\circled{4}$~holds due to the added entropy term $\sum P_{o} \log P_{o}$, which is a constant w.r.t. the model parameter~$\theta$. The standard posterior (Equation~\eqref{Standard_Posterior}) tends to zero under model misspecification ($P_o \neq P_ \theta$) as $N \rightarrow  +\infty$, while the approximate posterior (Equation~\eqref{extend_to_exp}) remains stable. This verifies our motivation for pursuing robust RA since vanilla RA, as well as data-augmentation based RAs, would inevitably output unreliably results even when infinity samples are available. 

\subsubsection{Types of distributional robustness and tractability} The choice of $D(\cdot, \cdot)$ in $\mathbb{P}(\theta | D(P_o, P_\theta) < \epsilon)$ (Equation~\eqref{theory_1}) affects both the richness of the robustness types as well as the tractability of the resulatnt optimization problem. The Wasserstein metric is a popular option in previous approaches on distributional robustness \citep{blanchet2016robust,gao2017wasserstein,volpi2018generalizing}, which exhibits superior tolerance to adversarially corrupted outliers \citep{chen2018outlier,chen2018robust} and also allows robustness to unseen data \citep{abadeh2015distributionally,sinha2017certifying}. Meanwhile, \citet{ben2013robust,namkoong2017variance} adopted $f$-divergences in pursuit of tractable optimization approaches. It is worthy noting that the rank aggregation task has its particularities. First, no generalization test is required for the RA task since we only need to aggregate the whole ranking dataset into one consensus full rank. Second, the probability ranking model itself has high complexity.  Therefore, we consider relative entropy for $D(\cdot, \cdot)$, since it allows standard inference with no additional computational burden and helps to exhibit robustness to most types of perturbations. 

Before introducing Theorem~\ref{extension_2}, we first introduce Lemma~1 which contains some preliminary results from \cite{miller2018robust}.

\begin{Lemma}[\cite{miller2018robust}]\label{Lemma_1}
	Let $\Delta_d = \{p\in\mathbb{R}^d: \sum^d_i p_i = 1, p_i > 0 \ \forall i\}$, and $q \in \Delta_d$. We argue that if $x_1, x_2, \ldots, x_N$ i.i.d. $\sim q$ and $F_N(t|x_{1:N}) = \frac{1}{N} \sum_{n=1}^N \mathbb{I}_{x_n}(t)$, then for $p \in \Delta_k$ near~$q$ in KL divergence,
	\begin{equation*}
	\mathbb{E}_{x_{1:N} \sim q} \left[\mathrm{exp}(-\alpha \mathcal{D}_{\mathrm{KL}}(p \| F_N(t|x_{1:N})))\right] \ \approx \ \left(\frac{N\tau_N}{\alpha}\right)^{ \frac{k-1}{2} }\mathrm{exp}(-N\tau_N \mathcal{D}_{\mathrm{KL}}(p \| q)),
	\end{equation*}
	where $\tau_N =  \frac{1/N}{1/N + 1/\alpha} $. \hfill\rule{2mm}{2mm}
\end{Lemma}

Lemma~\ref{Lemma_1} is defined for discrete distribution, which can be applied to our probability ranking model. To be specific, in terms of the item set~$\mathcal{O} = \{o_1, o_2, \ldots, o_M\}$, there are totally $\sum_{i=2}^M \tbinom{M}{i} i!$ possible ranking lists.  Therefore, the probability ranking model $P_ \theta$ is actually a discrete distribution with  $\sum_{i=2}^M \tbinom{M}{i} i!$  supports. 

\begin{Theorem}\label{extension_2}
Suppose relative entropy is adopted as the distance measure, namely $D(\mathcal{R}_N, \Re_N) = \mathcal{D}_{\mathrm{KL}}(F_N(x|\mathcal{R}_N)  \| F_N(x|\Re_N) )= \int F_N(x|\mathcal{R}_N)\mathrm{log}\frac{F_N(x|\mathcal{R}_N)}{F_N(x|\Re_N)}$, and the empirical distribution converges to the corresponding data generating distribution, namely $F_N(x|\mathcal{R}_N) \rightarrow P_o$ and $F_N(x|\Re_N) \rightarrow P_\theta$ when $N \rightarrow +\infty$. If $\epsilon$ is 
subject to an exponential prior, i.e., $\epsilon \sim \mathrm{Exp}(\alpha)$, we can obtain the following simple approximation to our CoarsenRank in Equation~\eqref{Coarsen_RA_Compact}:
\begin{equation}\label{extend_to_KL}
  \max_{\theta \in \Theta}\mathbb{E}_{\Re_N \sim P_\theta}\left[\mathbb{P}(\theta |  D(\mathcal{R}_N, \Re_N) < \epsilon)\right] \ \approxprop \  \max_{\theta \in \Theta}\pi(\theta)\prod_{n=1}^N P_{\theta}^{\tau_N}(\rho_n),
\end{equation}
where $\approxprop$ denotes that the term on the left is approximately equal to a term, which is proportional to the expression on the right, and $\tau_N = \frac{1/N}{1/N + 1/\alpha}$.
\end{Theorem}

\paragraph{Proof:} According to Theorem~\ref{extension_1}, we have 
\begin{equation*}\label{Bayesian_rule}
\begin{aligned}
\mathbb{E}_{\Re_N \sim P_\theta}\left[\mathbb{P}(\theta |  D(\mathcal{R}_N, \Re_N) < \epsilon)\right] \ &\propto \ \pi(\theta)\mathbb{E}_{\Re_N \sim P_\theta}\left[\mathbb{P}( D(\mathcal{R}_N, \Re_N) < \epsilon)| \theta)\right] \\
& = \ \pi(\theta)\mathbb{E}_{\Re_N \sim P_\theta}\left[\mathrm{exp}(-\alpha D(\mathcal{R}_N, \Re_N))\right],
\end{aligned}
\end{equation*}
where we omit the normalization constant with respect to $\theta$.

Further, we have 
\begin{equation*}
\begin{aligned}
\mathbb{E}_{\Re_N \sim P_\theta}\left[\mathrm{exp}(-\alpha D(\mathcal{R}_N, \Re_N))\right]
 & \overset{\circled{1}}{=} \mathbb{E}_{\Re_N \sim P_{\theta}}\left[ \mathrm{exp}(-\alpha \mathcal{D}_{\mathrm{KL}}(F_N(x|\mathcal{R}_N)  \| F_N(x|\Re_N)))\right] \\
&\overset{\circled{2}}{\approxprop} \mathrm{exp}(-N\tau_N \mathcal{D}_{\mathrm{KL}}(F_N(x|\mathcal{R}_N)  \| P_ \theta))\\
& \overset{\circled{3}}{\propto}  \mathrm{exp}\left(N\tau_N \sum_{n=1}^N F_N(\rho_n|\mathcal{R}_N) \mathrm{log} P_\theta(\rho_n)\right) \overset{\circled{4}}{ =}\prod_{n=1}^NP_{\theta}^{\tau_N}(\rho_n),
\end{aligned}
\end{equation*}
where $\tau_N =  \frac{1/N}{1/N + 1/\alpha} $.
$\circled{1}$ is valid by instantiating the distance measure with relative entropy.   $\circled{2}$ follows Lemma~\ref{Lemma_1} while omitting the constant-coefficient.
$\circled{3}$ holds due to the removal of the constant entropy term $\sum_{n=1}^N F_N(\rho_n|\mathcal{R}_N) \log F_N(\rho_n|\mathcal{R}_N)$, which is a constant w.r.t. the model parameter~$\theta$. $\circled{4}$ holds according to the definition of the empirical data distribution $F_N(x|\mathcal{R}_N) =  \frac{1}{N}  \sum_{n=1}^N  \mathbb{I}_{\rho_n}(x)$. \hfill\rule{2mm}{2mm}

\begin{Remark}[Connection between CoarsenRank and the standard posterior]
Since $\epsilon \sim \mathrm{Exp}(\alpha)$, we have $\mathbb{E}(\epsilon) = \frac{1}{\alpha}$ denoting the expected discrepancy of the collected preferences $\mathcal{R}_N$ w.r.t.~$\Re_N$. Further, $\mathbb{E}( \epsilon )$ tends to zero as $\alpha \rightarrow  + \infty$, which means the misspecification does not exist in the limit. Accordingly, the robust posterior Equation~\eqref{extend_to_KL} degenerates to the standard posterior as $\tau_N = \frac{1/N}{1/N + 1/\alpha}$ approximates to $1$ when $\alpha \rightarrow  + \infty$.  \hfill\rule{2mm}{2mm}
\end{Remark}

\subsection{Instantiating CoarsenRank with various probability ranking model}\label{Coarsened_Rank_model}
Based on the CoarsenRank framework (Equation~\eqref{extend_to_KL}), we instantiate $P_\theta$ with various probability ranking models.  

Please note an actual probability ranking model is usually defined over a subset of items and satisfies the summing up to 1 condition only on this subset. However, it does not necessarily satisfy the definition of the probability distribution over the whole item set $\mathcal{O}$, namely the probability integration of a probability ranking model over $\mathcal{O}$ is not $1$~\citep{zhao2019learning}\footnote{In terms of pairwise comparisons, there are totally $\tbinom{M}{2}$ distinct pairs with the corresponding probability integration being $\tbinom{M}{2}$. In terms of listwise preferences, there are $\sum_{i=2}^M \tbinom{M}{i}$ distinct subsets in total with the corresponding probability integration being $\sum_{i=2}^M \tbinom{M}{i}$.}. Without loss of generality, we assume each subset of items is uniformly sampled from the whole set. Then, the normalization constant can be safely omitted during optimization without affecting our final estimation. 


\subsubsection{Coarsened Thurstone model (CoarsenTH)}\label{Coarsen_Thurstone}

We first review the basic Thurstone model~\citep{thurstone1927method}, which is a popular ranking model to model pairwise comparisons.  Particularly, it assumes that the score $\theta_m \in \mathbb{R}$ for each item $o_m$ follows a Gaussian distribution $N(\mu_m, \sigma^2_m)$, $\forall m = 1, 2, \ldots, M$. For simplicity, we only consider the Thurstone model with $\sigma_m = 1$ for all items. In particular, the comparison between any two items $o_i$ and $o_j$ also follows a Gaussian distribution $N(\mu_i - \mu_j, 2)$. 

For a pairwise comparison $\rho_n: \rho_n^1> \rho_n^2$, Thurstone model assumes
\begin{equation}\label{TH_model}
P_{\theta}(\rho_n) = \Phi\left( \frac{ \triangle \theta_{\rho_n}}{\sqrt{2}}\right) = \frac{1}{\sqrt{2\pi}}\int_{-\infty}^{\frac{ \triangle \theta_{\rho_n}}{\sqrt{2}}} \text{exp}\left(-\frac{t^2}{2}\right)dt,
\end{equation}
where $ \triangle \theta_{\rho_n} = \theta_{\rho_n^1} - \theta_{\rho_n^2}$ denotes the difference between the score of the two items in $\rho_n$. $\Phi$ is the cumulative distribution function (CDF) of the standard normal distribution.

According to Theorem~\ref{extension_2}, an example of our CoarsenRank (Equation~\eqref{extend_to_KL}) using Thurstone model can be represented as follows:
\begin{eqnarray}\label{CoarsenTH}
\max_{\theta \in \Theta}\mathbb{E}_{\Re_N \sim P_\theta}\left[\mathbb{P}(\theta |  D(\mathcal{R}_N, \Re_N) < \epsilon)\right] & \approxprop & \max_{\theta \in \Theta}\pi(\theta)\prod_{n=1}^NP_{\theta}^{\tau_N}(\rho_n) \nonumber\\
&= & \max_{\theta \in \Theta}\frac{\pi(\theta)}{(\sqrt{2\pi})^{N\tau_N}}\left[\prod_{n=1}^N \int_{-\infty}^{\frac{ \triangle \theta_{\rho_n}}{\sqrt{2}}} \text{exp}\left(-\frac{t^2}{2}\right)dt\right]^{\tau_N},
\end{eqnarray}
where $\tau_N = \frac{1/N}{1/N + 1/\alpha}$. Equation~\eqref{CoarsenTH} can be only applied to pairwise comparisons.  When encountering listwise preferences, we need to split each listwise preferences into pairwise comparisons and then alternatively perform our CoarsenRank on the new dataset \citep{khetan2016data}.

\begin{Remark}[Optimization intractability and our strategy]\label{TH_Optimization}
The cumulative distribution function $\Phi$ is a special function, which cannot be expressed in terms of elementary functions. Therefore, it is inefficient or intractable to optimize Equation~\eqref{CoarsenTH} directly. 

Inspired by the work which explores the connection between the sigmoid function and the cumulative Gaussian distribution \citep{weng2011bayesian}, we consider approximating the cumulative distribution function $\Phi$ with the sigmoid function. In particular,
\begin{equation}\label{Gau2sigmoid}
\Phi\left(\frac{\triangle \theta_{\rho_n}}{ \sqrt{2}}\right)  \approx    \frac{1}{1 + \mathrm{exp}(-\lambda \triangle \theta_{\rho_n})}, 
\end{equation}
where $\lambda$ is set as $2/\sqrt{\pi}$ so that the two probability curves have the same slope at $\triangle \theta_{\rho_n} = 0$. A more accurate approximation with the second order moments guarantee can be found in \cite{daunizeau2017semi}. Then, Equation~\eqref{CoarsenTH} can be further approximated as 
\begin{equation}\label{CoarsenTH2}
\mathrm{Equation~\eqref{CoarsenTH}} \approx  \max_{\theta \in \Theta}\frac{\pi(\theta)}{(\sqrt{2\pi})^{N\tau_N}}\prod_{n=1}^N \frac{1}{\left[1 + \mathrm{exp}(-\lambda \triangle \theta_{\rho_n})\right]^{\tau_N}},
\end{equation}
where regular gradient-based optimization approaches could be carried out. In particular, the score is simply initialized to zero for all items, namely $\theta_m=0, \forall m = 1,2,\ldots, M$. 
\hfill\rule{2mm}{2mm}
\end{Remark}

\subsubsection{Coarsened Bradley-Terry model (CoarsenBT)}\label{Coarsen_BradlyTerry}

A closely related model to the Thurstone model is the Bradley-Terry (BT) model~\citep{bradley1952rank}. For any pairwise comparison $\rho_n: \rho_n^1> \rho_n^2$, the BT model assumes
\begin{equation}\label{BT_model}
P_{\theta}(\rho_n) = \frac{\theta_{\rho_n^1}}{\theta_{\rho_n^1}+\theta_{\rho_n^{2}}},
\end{equation}
where $\theta_m \in R_+^M$ is a positive support parameter for  item $o_m$, $\forall m = 1, 2, \ldots, M$.

According to Theorem~\ref{extension_2},an example of our CoarsenRank (Equation~\eqref{extend_to_KL}) using BT model can be represented as follows:
\begin{equation}\label{CoarsenBT}
\text{Equation~\eqref{extend_to_KL}}  \approxprop \max_{\theta \in \Theta}\pi(\theta)\prod_{n=1}^NP_{\theta}^{\tau_N}(\rho_n) = \max_{\theta \in \Theta}\pi(\theta)\prod_{n=1}^N \left[\frac{\theta_{\rho_n^1}}{\theta_{\rho_n^1}+\theta_{\rho_n^2}}\right]^{\tau_N},
\end{equation}
where $\tau_N = \frac{1/N}{1/N + 1/\alpha}$. Similar to Thurstone model, Equation~\eqref{CoarsenBT} can only model pairwise comparisons.  We still adopt the rank breaking strategy to split each listwise preferences into pairwise comparisons and perform our CoarsenRank on the new dataset \citep{khetan2016data}. 

\begin{Remark}[Optimization intractability and the data augmentation method]\label{Data_Augmentation}
The main inferential issue related to Equation~\eqref{CoarsenBT} concerns the presence of the annoying normalization terms $\theta_{\rho_n^1}+\theta_{\rho_n^2}$, $\forall n = 1,2,\ldots, N$, that do not permit the direct maximization of the posterior. Further, the nonnegative constraint over the model parameters $\theta$ rules out the direct applications of gradient-based optimization approaches.  

Motivated by~\cite{caron2012efficient}, we introduce the data augmentation method to address the above-mentioned difficulty. Considering the fact that the Gumbel distribution is employed as a distribution of the support parameters and the conjugacy of the Gamma density with the Gumbel distribution,  we follow~\cite{caron2012efficient} and introduce an auxiliary Gamma random variable for each normalization term, which leads to a joint distribution without suffering from the annoying normalization terms.   \hfill\rule{2mm}{2mm}
\end{Remark}

\subsubsection{Coarsened Plackett-Luce model (CoarsenPL)}\label{CoarsenPL}
Here we instantiate $P_{\theta}$ with the popular Plackett-Luce (PL) model  \citep{plackett1975analysis,luce1959individual}. Different from the previous Thurstone mode and the BT model, PL model is a more general probability ranking model, which could model listwise rankings of a finite set of items directly. Note that PL model incorporates BT model as a special~case. 

For a ranking list $\rho_n: \rho_n^1> \rho_n^2> \ldots> \rho_n^k$, the PL model assumes
\begin{equation}\label{PL_model}
P_{\theta}(\rho_n) = \prod_{i=1}^{k-1}\frac{\theta_{\rho_n^i}}{\theta_{\rho_n^i}+\theta_{\rho_n^{i+1}}+\ldots + \theta_{\rho_n^k}},
\end{equation}
where $\theta_m \in R_+^M$ is a positive support parameter associated with item $o_m$, $\forall m = 1, 2, \ldots, M$. Comparing Equation~\eqref{PL_model} to Equation~\eqref{BT_model}, PL model degenerates to BT model when modeling pairwise comparisons, i.e., $k \equiv 2$.

According to Theorem~\ref{extension_2}, an example of our CoarsenRank (Equation~\eqref{extend_to_KL}) instantiated using PL model can be represented as follows:
\begin{equation}\label{CoarsenRank}
\begin{aligned}
\text{Equation~\eqref{extend_to_KL}} \ \approxprop \ \max_{\theta \in \Theta}\pi(\theta)\prod_{n=1}^NP_{\theta}^{\tau_N}(\rho_n) = \max_{\theta \in \Theta}\pi(\theta)\prod_{n=1}^N \left[\prod_{i=1}^{k-1}\frac{\theta_{\rho_n^i}}{\theta_{\rho_n^i}+\theta_{\rho_n^{i+1}}+\ldots + \theta_{\rho_n^k}}\right]^{\tau_N}.
\end{aligned}
\end{equation}
where $\tau_N = \frac{1/N}{1/N + 1/\alpha}$. $k$ denotes the length of each preference, which could be variant for different preferences. 

\begin{Remark}[Optimization intractability and the data augmentation method]\label{Data_Augmentation_PL}
The main inferential issue related to Equation~\eqref{CoarsenRank} concerns the presence of the annoying normalization terms $\theta_{\rho_n^i}+\theta_{\rho_n^{i+1}}+\ldots + \theta_{\rho_n^k}$, $\forall n = 1,2,\ldots, N, i = 1,2,\ldots, k-1$, that do not permit the direct maximization of the posterior. Further, the nonnegative constraint over the model parameters $\theta$ rules out the direct applications of gradient-based optimization approaches.  

Following our analysis in Remark~\ref{Data_Augmentation}, we avoid this issue using the data augmentation method. In particular, we introduce an auxiliary Gamma random variable for each normalization term $\theta_{\rho_n^i}+\theta_{\rho_n^{i+1}}+\ldots + \theta_{\rho_n^k}$, $\forall n = 1,2,\ldots, N, i = 1,2,\ldots, k-1$. Then,  the resultant joint distribution would no longer suffers from the annoying normalization terms.  \hfill\rule{2mm}{2mm}
\end{Remark}

\subsection{Connection between CoarsenRank and Mallows model}\label{Mallows_extension}

Permutation-based models are based on the definition of distances $D(\cdot, \cdot)$, which express the distance of a permutation to the ground truth permutation. The most prominent example of these models is the Mallows model (MM)~\citep{mallows1957non}, an exponential model that expresses the probability of a permutation in terms of its distance to a reference permutation. Let $\rho$ be a full/partial ranking list, then MM specifies:  
\begin{equation}\label{Mallows}
P(\rho| r)= \frac{1}{\psi(\alpha)} \exp(-\alpha D(\rho, r)), \quad \text { where } \psi(\alpha)=\sum_{\rho} \exp (-\alpha D(\rho, r)).
\end{equation}
Here $\alpha \in \mathbb{R}_+$ is a spread parameter and $r$ is the reference permutation, or called the unknown ground truth. $D(\rho, r)$ represents a sample-level distance between $\rho$ and $r$. Note that $r$ is the mode, and the closer a ranking list $\rho$ is to $r$, the larger $p(\rho)$ is. The alternative distance measures considered are Kendall tau distance, Spearman's rank distance, etc~\citep{diaconis1988group}. 

According to Equation~\eqref{Mallows}, the probability of MM for $\mathcal{R}_N$ could be represented~as
\begin{equation}\label{Mallow_Full}
\begin{aligned}
P(\mathcal{R}_N| r)  &= \frac{1}{\psi^N(\alpha)}\prod_{n =1}^N \exp(-\alpha D(\rho_n, r)) = \frac{1}{\psi^N(\alpha)}\exp(-\alpha \sum_{n =1}^ND(\rho_n, r))\\
&\overset{\circled{1}}{=}\frac{1}{\psi^N(\alpha)}\exp(-\alpha D(\mathcal{R}_N, r)) \overset{\circled{2}}{\propto} \exp(-\alpha D(\mathcal{R}_N, r)),
\end{aligned}
\end{equation}
where $\mathcal{R}_N = \{\rho_1, \rho_2, \ldots, \rho_N\}$ denotes the collected preferences. $\circled{1}$ is valid since we define $D(\mathcal{R}_N, r) = \sum_{n =1}^ND(\rho_n, r)$. $\circled{2}$ holds due to the omission of the data non-relevant normalization term.

However, permutation-based models are often impractical for the large-scale problem, because: (1) the normalization term $\psi(\alpha)$ usually requires high computational cost due to discrete distance computation; and (2) a maximum likelihood estimation involves an impossible discrete search for ranking over a large volume of items. 

\begin{Remark}[\label{Connection_to_mallow} Comparison between CoarsenRank and MM] Comparing 
Equation~\eqref{Mallow_Full} to our CoarsenRank formulation~(Equation~\eqref{extend_to_exp}), we can find that the difference between CoarsenRank and MM mainly lies in the definition of distance $D(\cdot, \cdot)$. Namely, distribution-level distance, i.e., $\mathcal{D}_{\mathrm{KL}}(\cdot, \cdot)$ is adopted for CoarsenRank, while sample-level distance, e.g., Kendall tau distance, is adopted for MM. The superiority of CoarsenRank over MM lies in three aspects: (1) In terms of inference, an efficient inference strategy is discussed in the Remark after each variant of CoarsenRank, respectively;  (2) In terms of explanation, our CoarsenRank formulation is derived from the Coarsening mechanism while assigning an exponential prior for the size of the neighborhood; (3) In terms of  optimization w.r.t. the hyperparameter $\alpha$, we avoid parameter turning by adopting data-driving strategy for choosing $\alpha$, while Mallow’s mode does not enjoy this convenience.
\end{Remark}

\section{Efficient Bayesian inference}\label{Sect_4}
Following the discussion in the previous section, we propose two  equivalent algorithms to solve the proposed  CoarsenRank framework efficiently, namely a closed-form Expectation-Maximization algorithm in Section~\ref{EM} and a Gibbs Sampling algorithm in Section~\ref{Gibbs}. Further, we discuss their algorithm complexity in Section~\ref{time-space}.  To avoid hyperparameter tuning, an efficient model selection method based on Algorithm~\ref{CoarsenRank_Algo_Gibbs} is introduced in Section~\ref{DIC} to choose the single hyperparameter in a data-driven manner.

In particular,  we focus on Coarsened PL model (Equation~\eqref{CoarsenRank}) only and refer it as CoarsenRank for two reasons:  (1) PL model can be applied to preferences with various length and incorporates BT model as a special case; (2) Thurstone model and BT model are constrained to pairwise preferences, while the rank breaking strategy would lead to computational inefficient.

\subsection{Data augmentation method for eliminating the normalization terms}\label{Data_Augment}

First, we reformulate Equation~\eqref{CoarsenRank} as follows:
\begin{align}\label{CoarsenRank_2}
	\text{Equation~\eqref{CoarsenRank}} &= \max_{\theta \in \Theta}\pi(\theta)\prod_{n=1}^N \left[\prod_{i=1}^{k-1}\frac{\theta_{\rho_n^i}}{\theta_{\rho_n^i}+\theta_{\rho_n^{i+1}}+\ldots + \theta_{\rho_n^k}}\right]^{\tau_N} \\
	&= \max_{\theta \in \Theta}\pi(\theta)\prod_{n=1}^N\prod_{i=1}^{k-1}\left[ \frac{\theta_{\rho_n^i}}{\theta_{\rho_n^i}+\theta_{\rho_n^{i+1}}+\ldots + \theta_{\rho_n^k}}\right]^{\tau_N} 
	=\max_{\theta \in \Theta}\pi(\theta)\prod_{n=1}^N\prod_{i=1}^{k-1}\left(\frac{\theta_{\rho_n^i}}{\eta_n^i} \right)^{\tau_N},\nonumber
\end{align}    
where $\eta_n^i = \theta_{\rho_n^i}+\theta_{\rho_n^{i+1}}+\ldots + \theta_{\rho_n^k}$, $\forall n = 1,2,\ldots, N, i = 1,2,\ldots, k-1$. Due to the presence of the normalization terms $\eta_n^i$ in Equation~\eqref{CoarsenRank_2}, the direct maximization of the posterior w.r.t. to the nonnegative parameter $\theta$ would encounter significant inefficiency. 

For example, if we want to eliminate the term $\eta_n^i$ appearing in the denominator,  we could introduce an auxiliary random variable, whose conditional distribution contains $\eta_n^i$ in the numerator. Then, the resultant joint distribution would no longer contain this annoying $\eta_n^i$ in the denominator. In particular, one particular Gamma distribution is a promising candidate that satisfies the above requirements, i.e., $\text{Gam}(\xi| 1, q) = qe^{-q\xi}$ could be used to eliminate a $q$ containing in the denominator.

Therefore, we introduce an auxiliary variable $\xi_n^i$ with regarding to each $\eta_n^i$, $\forall n = 1,2, \ldots, N$ and $\forall i = 1,2, \ldots, k-1$. Further, we define the posterior distribution of $\xi_n^i$ as follows,
\begin{equation}\label{data_posterior}
P(\xi_n^i | \rho_n, \theta) = \text{Gam}(\xi_n^i| 1, \eta_n^i) = \eta_n^i e^{-\xi_n^i\eta_n^i}.
\end{equation}
Then, we can deal with the joint distribution directly, i.e., 
\begin{equation}\label{joint_distribution}
P(\rho_n, \Xi | \theta) = P_{\theta}(\rho_n) \prod_{i=1}^{k-1}P(\xi_n^i | \rho_n, \theta) = \prod_{i=1}^{k-1}\left(\frac{\theta_{\rho_n^i}}{\eta_n^i} \cdot \eta_n^i e^{-\xi_n^i\eta_n^i}\right)= \prod_{i=1}^{k-1}\left(\theta_{\rho_n^i} e^{-\xi_n^i\eta_n^i}\right),
\end{equation}
where $\Xi = \{\xi_n^i\}_{i=1}^{k-1}$ denotes the introduced auxiliary variables.

Further, we utilize a Gamma prior to instantiate the prior distribution $\pi(\theta)$, which naturally satisfies the nonnegative constraint of $\theta$, i.e., $\theta \sim \text{Gam}(\theta| a, b) =  \prod_{m=1}^M\text{Gam}(\theta_m| a_m, b_m)$). Therefore, the full likelihood of our CoarsenRank model (Equation~\eqref{CoarsenRank_2}) can be formulated as follows, 
\begin{align}\label{MAP_CoarsenRank}
    P(\mathcal{R}_N, \Xi, \theta, \epsilon | \{a_m, b_m\}_{m=1}^M, \alpha) &= \pi(\theta)\prod_{n=1}^N \Big(P(\rho_n, \Xi | \theta)\Big)^{\tau_N} \\
    & = \prod_{m=1}^M\text{Gam}(\theta_m| a_m, b_m) \prod_{n=1}^N\prod_{i=1}^{k-1}\left(\theta_{\rho_n^i} e^{-\xi_n^i\eta_n^i}\right)^{\tau_N}, \nonumber
\end{align}
where $\tau_N = \frac{1/N}{1/N + 1/\alpha}$. $\mathcal{R}_N = \{\rho_1, \rho_2, \ldots, \rho_n\}$ denotes the observed preferences. $\epsilon \sim \mathrm{Exp}(\alpha)$ is the discrepancy between the collected preferences $\mathcal{R}_N$ and its idealized counterpart $\Re_N$, measured in relative entropy. $(a_m, b_m)$ is initialized to (1, 2), $\forall m = 1, 2, . . ., M$. We fixed $\{a_m, b_m\}_{m=1}^M$ in this paper to eliminate their coupling effects with other factors in CoarsenRank~(Equation~\eqref{MAP_CoarsenRank}). 

\subsection{EM algorithm with closed-formed updating rules}\label{EM}

Concerning the presence of the introduced auxiliary variables $\Xi$, we resort to the Expectation-Maximization (EM) framework, which is a silver bullet to compute the maximum-likelihood solution or maximum a posterior estimation in the presence of latent variables.

\begin{algorithm}[!tb]
\caption{ Closed form EM for Coarsened rank aggregation (CoarsenRank)\label{CoarsenRank_Algo}}
\begin{algorithmic}[1]
\STATE \textbf{Input:} the collection of preferences $\mathcal{R}_N$, the number of iteration $T$.
\STATE \textbf{Initialization:} hyperparameters$\{a_m, b_m\}_{m=1}^M$ for $\theta$.
\FOR{$t = 1,2,\dotsc,T$}
\STATE \textit{E-step:} calculate the posterior expectation of auxiliary variable $\xi_n^i$ according to Equation~\eqref{posterior}.\
\STATE \textit{M-step:} update $\theta_m$ according to Equation~\eqref{theta_update} $\forall m = 1,2,\ldots M$.\
\ENDFOR
\STATE \textbf{Rank:} the item score $\theta$ to derive the final ranking.
\STATE \textbf{Output:} the final ranking.
\end{algorithmic}
\end{algorithm}

\paragraph{Expectation step (E-step)} In the expectation step, we calculate the expectation of each auxiliary variable $\xi_n^i$ w.r.t. its posterior distribution $P(\xi_n^i | \rho_n, \theta)$:
\begin{equation}\label{posterior}
\mathbb{E}_{P(\xi_n^i | \rho_n, \theta)}[\xi_n^i] = \frac{1}{\theta_{\rho_n^i}+\theta_{\rho_n^{i+1}}+\ldots + \theta_{\rho_n^k}} = \frac{1}{\eta_n^i},
\end{equation}
where $n = 1,2, \ldots, N$ and $i = 1,2, \ldots, k-1$. Then, the expectation of the complete-data log-likelihood function w.r.t. the posterior of the introduced auxiliary variables $\Xi$ can be represented as follows:
\begin{align}\label{log_likelihood}
&\mathbb{E}_{P(\Xi | \mathcal{R}_N, \theta)} \left[\text{log} P(\mathcal{R}_N,  \Xi, \theta, \epsilon | \{a_m, b_m\}_{m=1}^M, \alpha)\right]\\
&\quad = \sum_{m=1}^M\big[(a_m-1)\log \theta_m - b_m\theta_m\big]  + \sum_{n=1}^N\sum_{i=1}^{k-1}\big[\tau_N \log\theta_{\rho_n^i} - \tau_N \mathbb{E}[\xi_n^i]\eta_n^i\big]+ \text{constant}, \nonumber
\end{align}
where $P(\Xi | \mathcal{R}_N, \theta) = \prod_{n=1}^N\prod_{i=1}^{k-1} P(\xi_n^i |\rho_n, \theta)$ following Equation~\eqref{posterior}.

\paragraph{Maximization step (M-step)} In the maximization step, we maximize the objective function Equation~\eqref{log_likelihood} by setting its gradient w.r.t. $\theta_m$~to zero and obtain the following estimates for $\theta_m$ $\forall m= 1,2,\ldots, M$:
\begin{equation}\label{theta_update}
\theta_m =  \frac{\tau_N\sum_{n=1}^N\sum_{i=1}^{k-1}(\varphi_{n,i}^m) + a_m-1}{\tau_N\sum_{n=1}^N\sum_{i=1}^{k-1}(\psi_{n,i}^m \cdot \mathbb{E}[\xi_n^i]) + b_m}, 
\end{equation}
where $\varphi_{n,i}^m =\begin{cases}1 & \rho_n^i = m\\0 & \text{otherwise}\end{cases}$ and $ \psi_{n,i}^m =\begin{cases}1 & m \in \{\rho_n^i, \ldots, \rho_n^k\}\\0 & \text{otherwise}\end{cases}.$ 

Overall, the EM algorithm for Coarsened rank aggregation (CoarsenRank) is summarized in Algorithm~\ref{CoarsenRank_Algo}.

\subsection{Gibbs sampling for Coarsened rank aggregation}\label{Gibbs}

In our CoarsenRank (Equation~\eqref{MAP_CoarsenRank}), there are two types of latent variables, i.e., $\Xi$ and~$\theta$. According to our definition, the posterior distribution of $\xi_n^i$ can be represented as
\begin{equation}\label{Xi}
P(\xi_n^i | \rho_n, \theta) = \text{Gam}(\xi_n^i| 1, \eta_n^i) = \eta_n^i e^{-\xi_n^i\eta_n^i},
\end{equation}
where $n = 1, 2, \ldots, N$ and $i = 1,2, \ldots, k$. Similarly, the full conditional distributions of $\theta_m$ $\forall m = 1, 2, \ldots, M$ are still members of the Gamma family. According to Equation~\eqref{theta_update}, the posterior distribution $P(\theta_m | \mathcal{R}_N, \Xi)$ can be represented as
\begin{align}\label{theta}
P(\theta_m | \mathcal{R}_N, \Xi) = \text{Gam}\left(\theta_m \mid  \tau_N\sum_{n=1}^N\sum_{i=1}^{k-1}(\varphi_{n,i}^m) + a_m-1, \tau_N\sum_{n=1}^N\sum_{i=1}^{k-1}(\psi_{n,i}^m \cdot \mathbb{E}[\xi_n^i]) + b_m\right),
\end{align}
where $\varphi_{n,i}^m =\begin{cases}1 & \rho_n^i = m\\0 & \text{otherwise}\end{cases}$ and $ \psi_{n,i}^m =\begin{cases}1 & m \in \{\rho_n^i, \ldots, \rho_n^k\}\\0 & \text{otherwise}\end{cases}.$ 

Therefore, the Gibbs sampling procedure for Coarsened rank aggregation (CoarsenRank) can be summarized in Algorithm~\ref{CoarsenRank_Algo_Gibbs}.

\begin{algorithm}[!tp]
\caption{Gibbs Sampling for Coarsened rank aggregation (CoarsenRank)\label{CoarsenRank_Algo_Gibbs}}
\begin{algorithmic}[1]
\STATE \textbf{Input:} the collection of preferences $\mathcal{R}_N$, the number of iteration $T$.
\STATE \textbf{Initialization:} hyperparameters $\{a_m, b_m\}_{m=1}^M$ for $\theta$, burn-in iterations $\Delta = 100$.
\FOR{$t = 1,2,\dotsc,T+\Delta$}
\STATE sample $\xi_n^i$ from $P(\xi_n^i | \rho_n, \theta^{t-1})$ according to Equation~\eqref{Xi}, $\forall n = 1, 2, \ldots, N$, $\forall i = 1,2, \ldots, k$.\
\STATE sample $\theta^t_m$ from $P(\theta_m | \mathcal{R}_N, \Xi)$ according to Equation~\eqref{theta}, $\forall m = 1, 2, \ldots, M$.\
\ENDFOR
\STATE \textbf{Average:} the resulting $\theta^{\Delta+1:\Delta+T}$ to $\bar{\theta}$.
\STATE \textbf{Rank:} the item score $\bar{\theta}$ to derive the final ranking.
\STATE \textbf{Output:} the final ranking.
\end{algorithmic}
\end{algorithm}

\subsection{Time and space complexity analysis}\label{time-space}

We analyze the algorithm complexity for our CoarsenRank as well as vanilla RA methods to further demonstrate the superiority of our CoarsenRank. Note that CoarsenBT and CoarsenPL denote the optimization strategy in Algorithm~\ref{CoarsenRank_Algo}, while CoarsenPL(GS) denotes the optimization strategy in Algorithm~\ref{CoarsenRank_Algo_Gibbs}.

Let $N$, $M$, $T_1$, $T_2$, and $T_3$ denote the number of preferences, the number of items, the number of gradient descent steps, the number of EM iterations, and the number of Gibbs sampling iterations, respectively. For ease of analysis, we assume the length of all preferences equals $k$. The time and space complexities of CoarsenRank are listed in Table~\ref{time_complexity}.

\begin{table}[!ht]
\centering
\caption{\label{time_complexity} Complexity analysis of CoarsenRank}
\renewcommand{\arraystretch}{1.2}
\setlength{\tabcolsep}{1.2mm}{	
\scalebox{1}{
\begin{tabular}{c|c|ccc}\toprule[1.3pt]
\multicolumn{2}{c|}{Complexity}  & Optimization & Time   & Space \\\midrule[1pt]
\multirow{4}{*}{Vanilla} 
& TH  & Gradient descent & $\mathcal{O}(NMT_1k^2)$ & $\mathcal{O}(NMk^2)$      \\
& BT  & Expectation maximization & $\mathcal{O}(NMT_2k^2)$&$\mathcal{O}(NMk^2)$ \\
& PL-EM& Expectation maximization& $\mathcal{O}(NMT_2k^2)$&$\mathcal{O}(NMk)$ \\
& PL  & Gradient descent & $\mathcal{O}(NMT_1k^2)$  & $\mathcal{O}(NMk)$\\\hline
\multirow{2}{*}{Robust}  
& CrowdBT & Bayesian moment matching &  $\mathcal{O}(NT_1k^2)$   & $\mathcal{O}(\max(M,G))$   \\
& PeerGrader& Gradient descent & $\mathcal{O}(NMT_1k^2)$ & $\mathcal{O}(NMk)$\\ \hline
\multirow{4}{*}{Coarsen} 
& CoarsenTH & Gradient descent  & $\mathcal{O}(NMT_1k^2)$& $\mathcal{O}(NMk^2)$ \\
& CoarsenBT & Expectation maximization & $\mathcal{O}(NMT_2k^2)$& $\mathcal{O}(NMk^2)$\\
& CoarsenPL & Expectation maximization & $\mathcal{O}(NMT_2k^2)$& $\mathcal{O}(NMk)$\\
& CoarsenPL(GS)& Gibbs sampling & $\mathcal{O}(NM(T_3+\Delta)k^2)$ & $\mathcal{O}(NMk + MT_3)$\\\bottomrule[1.3pt]
\end{tabular}}}
\end{table}
Let's give more discussions about Algorithm~\ref{CoarsenRank_Algo} and Algorithm~\ref{CoarsenRank_Algo_Gibbs} in terms of  time complexity:
\begin{enumerate}
\item Our CoarsenRank consists of $T$ iterations between Equation~\eqref{posterior} and Equation~\eqref{theta_update}. The main computational routine is the denominator $\sum_{n=1}^N\sum_{i=1}^{k-1}(\psi_{n,i}^m \cdot \mathbb{E}[\xi_n^i])$ for $m = 1,2, \ldots, M$, whose computational cost is $\mathcal{O}(NMk^2)$ per iteration. The same computations also apply to PL-EM since the only difference between PL-EM and CoarsenPL is the value of the scalar  $\tau_N$, which 
refers to PL-EM if $\tau_N=1$, and CoarsenPL if $0<\tau_N<1$.

\item Pairwise RA methods, e.g., CoarsenTH and CoarsenBT, need to split each $k$-ary preference into $\frac{k(k-1)}{2}$ pairwise preferences, thus increasing the number of preferences from $N$ to $\frac{Nk(k-1)}{2}$, but shorten the computation cost per preference from $\mathcal{O}(\frac{k(k-1)}{2})$  to $\mathcal{O}(1)$. The analysis also apply to TH and BT since the only difference between CoarsenRank (CoarsenTH and CoarsenBT) and vanilla RA (TH and BT) is the value of the scalar  $\tau_N$.

\item As for each iteration, gradient descent-based optimization (TH, CoarsenTH and PL) has the same computational cost, i.e., $\mathcal{O}(NMk^2)$  as that of EM-based ones (BT, CoarsenBT, PL-EM and CoarsenPL), since it needs to calculate the gradient of $M$ items over $\frac{Nk(k-1)}{2}$ pairwise preferences for BT-based models or $N$ $k$-ary preferences for PL-based models.

\item CrowdBT adopts Bayesian moment matching for online updates. Although with complex updating rules, it only updates the emerging item in the current preference without traversing all items. Namely, its computation cost is not relevant to $M$. 

\item PeerGrader introduces an extra variable to model the grader reliability. Since it adopts gradient descent for optimization, PeerGrader has the same time complexity as PL. However, an iterative alternating-minimization is required for optimizing the item score and the grader reliability, which incurs more computations than vanilla PL. 

\item The analysis for CoarsenPL in Algorithm 1 also applies to CoarsenPL(GS) of Algorithm 2 by comparing Eq.31 and Eq.29, which, however, requires more iterations (i.e., $T_3 \gg T_2$). Furthermore, CoarsenPL(GS) also includes an extra computation cost $\mathcal{O}(NM\Delta k^2)$ for the burn-in process. 
\end{enumerate}

In terms of space complexity: 
\begin{enumerate}
    \item The primary storage elements are the index $\varphi_{n,i}^m$ and $\psi_{n,i}^m$, where $n =1,2,\ldots, N$, $i=1,2,\ldots, k$ and $m =1,2,\ldots, M$. Therefore, the overall space complexity for PL-based methods, e.g., PL, PL-EM and CoarsenPL, is $\mathcal{O}(NMk)$. 
    
    \item Apart from the same storage as other PL-based methods, PeerGarder needs to store the reliability (a scalar) for each grader as well, which however can be omitted compared to the storage of indies. Therefor, the space complexity for PeerGarder is also $\mathcal{O}(NMk)$. 
    
    \item Pairwise RA methods, e.g., TH, CoarsenTH and BT, need to split each $k$-ary preference into $\frac{k(k-1)}{2}$ pairwise preferences, thus increasing the number of preferences from $N$ to $\frac{Nk(k-1)}{2}$, so the overall space complexity is $\mathcal{O}(NMk^2)$. 
    
    \item Since CrowdBT adopts an online updating paradigm, it only needs to store a two-dimensional hyperparameter for each score, a two-dimensional hyperparameter for each grader. Let $G$ denote the number of graders, the over space complexity for CrowdBT is $\mathcal{O}(\max(M,G))$.
    
    \item Similarly, apart from the storage of the two index matrices $\varphi$ and $\psi$, CoarsenPL(GS) requires extra storage about $MT_3$ for storing $T_3$ intermediate variables ($\theta \times T_3$) so as to calculate the final item score.
\end{enumerate}

Note that each item only appears in a small number of preferences, which means there is no need to store all $N$ preferences for each item, especially when the length of preference ($k$) is small. Therefore,  we can choose a more efficient strategy for storing the preferences, namely, regarding each item, we only store the relevant preferences that contain the target item. Assume the maximum number of relevant preferences is $\tilde{N}\ll N$, which can reduce both the time complexity and space complexity of CoarsenPL from $\mathcal{O}(NMT_2k^2)$ and $\mathcal{O}(NMk)$ to $\mathcal{O}(\tilde{N}MT_2k^2)$ and $\mathcal{O}(\tilde{N}Mk)$, respectively. The same is true when applying to other RA methods. We adopt the proposed strategy for all RA methods in the experiment. 

\subsection{A data-driven strategy for choosing $\alpha$}\label{DIC}

Regarding the hyperparameter optimization, we focus on exploring the effects of the hyperparameter $\alpha$ in this paper. Other hyperparameters were not further explored in our experiment since these are not our focus in this paper.  

Regarding the hyperparameter $\alpha$, we have no prior basis for choosing parameter $\alpha$ in Equation~\eqref{extend_to_exp}. Therefore, the following diagnostic curve can help to make a data-driven choice. Let $f(\alpha)$ be a measure of fit to the data and $g(\alpha)$ be a measure of model complexity. Following \citep{spiegelhalter1998bayesian}, we use the posterior expected log-likelihood for $f(\alpha)$, and the difference between the log-likelihood evaluated at the posterior mean of the parameters and the posterior expected log-likelihood for $g(\alpha)$. Specifically, we~define 
\begin{equation}
 f(\alpha) = \mathbb{E}_{q_{\alpha}(\theta|\mathcal{R}_N)}\left[\mathrm{log}P_{\theta}(\mathcal{R}_N)\right] \quad \text{and} \quad g(\alpha) = \mathrm{log}P_{\mathbb{E}(\theta)}(\mathcal{R}_N) - f(\alpha), 
\end{equation}
where $q_{\alpha}(\theta|\mathcal{R}_N)$ is an approximate posterior distribution for $\theta$, i.e., Equation~\eqref{theta}, and $\mathbb{E}(\theta) = \mathbb{E}_{q_{\alpha}(\theta|\mathcal{R}_N)}[\theta]$ is the posterior expectation of $\theta$. 

Therefore, the adopted Deviance Information Criterion (DIC) \citep{spiegelhalter1998bayesian} can be represented as
\begin{equation}\label{DIC_eq}
    \mathrm{DIC} = g(\alpha) - f(\alpha).
\end{equation}
As $\alpha$ ranges from $0$ to $+\infty$, DIC traces out a curve in $\mathbb{R}$, and the technique is to choose $\alpha$ with the lowest DIC or where DIC levels off.

\section{Perturbation assumption and rank model assumption}\label{related-work}
In this section, we summarize the differences between our CoarsenRank and related RA models in terms of distance measure, perturbation assumption and ranking model assumption. A detailed comparison is listed in Table~\ref{noisy_assumption}. 

\begin{table}[!t]
\large
	\centering
	\caption{\label{noisy_assumption} Comparison between various ranking models in terms of their distance measure, perturbation assumption and model assumption. ``---'' denotes the assumption does not apply to a particular ranking model.}
	\renewcommand{\arraystretch}{1.2}
	\setlength{\tabcolsep}{1.2mm}{	
	\scalebox{0.7}{
	\begin{tabular}{c|c|c|cc|ccc}\toprule[1.3pt]
\multicolumn{2}{c|}{\multirow{2}{*} {\bf{Baselines}} } & \bf{Distance measure}  & \multicolumn{2}{c|}{\bf{Perturbation assumption}}       & \multicolumn{3}{c}{\bf{Ranking model assumption}}\\\cline{3-8}
\multicolumn{2}{c|}{}   & \makecell{distribution or\\ sample level} & \makecell{perturbation \\$P(\rho_n|\varrho_n )$} & \makecell{neighborhood \\size $\epsilon$} &  \makecell{model \\$P_{\theta}$} &  \makecell{prior \\$\pi(\theta)$} &  \makecell{posterior \\ $P(\xi_n^i | \rho_n, \theta)$} \\\toprule[1.3pt]
\multirow{3}{*}{Vanilla}   & TH    & distribution  &       ---       &       ---           & TH  &   Gaussian    &    Gaussian \\ 
       & BT    & distribution  &       ---       &       ---           &           BT  &   Gaussian/Gamma   &    same as prior \\
       &PL   & distribution  &       ---       &       ---           & PL  &   Gaussian/Gamma   &    same as prior \\\hline
\multirow{4}{*}{Robust}  & MM    &  sample  &  fractional likelihood            &    ---      &     MM              &          ---          &     ---                     \\
& CrowdBT   & distribution                  &     Dawid-Skene model &          ---         &         BT   &    Gaussian       &      Gaussian                    \\
        & ROPAL   & distribution                  &     Dawid-Skene model &          ---         &         PL   &    Gaussian       &      Gaussian                    \\
                & PeerGrader       &  distribution                 &    fractional likelihood   & ---  &   BT/PL/TH   &    Gaussian    &       ---                   \\\hline
\multirow{3}{*}{Coarsen}   & CoarsenTH        &  distribution                &   ---              &    exponential prior                   & TH  &   Gaussian    &    ---  \\
    & CoarsenBT        &  distribution          &   ---              &    exponential prior          &      BT  &   Gamma   &    Gamma                   \\
    & CoarsenPL        &   distribution                &   ---              &    exponential prior                  & PL  &   Gamma   &    Gamma \\
\toprule[1.3pt]\end{tabular}}}
\end{table}

Specifically, we classify the rank aggregation methods into three categories: Vanilla RA, Robust RA, and Coarsen RA. Vanilla RA, such as Thurstone (TH) model~\citep{thurstone1927method}
Bradley-Terry (BT) model~\citep{bradley1952rank}
Plackett-Luce (PL) model~\citep{plackett1975analysis,luce1959individual}, assumes all ranking lists are generated from the same distribution, under some parameterized ranking model, e.g., TH, BT, and PL. Further, a prior is usually introduced for the parameter to avoid overfitting. 

The formulation of MM is similar to the fractional likelihood~\citep{bhattacharya2019bayesian} where the tunable spread parameter in MM has the same function as the exponential variable in a fractional likelihood. This is why we classify MM as a robust RA. See section~\ref{Mallows_extension} for more detailed discussion.

Robust RA extends vanilla RA by considering extra noisy perturbations incurred during the data collection. Representative robust RAs are CrowdBT~\citep{chen2013pairwise} and 
ROPAL~\citep{han2018robust}, which recovers the idealized ranking lists by some predefined perturbation mechanisms, following the Dawid-Skene model. Meanwhile, PeerGrader~ \citep{raman2014methods} also enhances the robustness of vanilla RA following the principle of fractional likelihood. In particular, PeerGrader introduces an extra parameter for each user, which would decrease the effects of noisy preferences from an unreliable user. Above Robust RAs aim at capturing a perturbation pattern w.r.t. to each user, which requires sufficient available preferences from each user. This constraint is too strong while only one preference from each user is usually available in real application. 

Our CoarsenRank is motivated by the Coarsening mechanism, where we perform regular RA over the neighborhood of original ranking data. We only introduce one extra parameter, representing the size of the neighborhood. A simple formulation of CoarsenRank is further derived if we adopt relative entropy as the distance measure and assign an exponential prior for the unknown neighborhood size. Three variants of CoarsenRank are introduced by instantiating with the vanilla TH, BT, and PL model, and an efficient EM algorithm could be derived under the same assumption as the vanilla RAs.

\section{Experimental evaluation}\label{experiment}

In this section, we verify the efficacy of the proposed CoarsenRank algorithm on noisy rank aggregation with the state-of-the-art approaches. The results are carried on four real-world noisy ranking datasets.

\subsection{ Experimental setting}\label{experiment_setting}

\paragraph{Performance metric:} Regarding the performance metric, we consider the Kendall tau similarity \citep{kendall1938new}, which is one of the most common measures of similarity between rankings, namely  
\begin{equation}\label{acc_metric}
\tau( \sigma_1 ,  \sigma_2) = \frac{1}{\overline{M}} \sum_{i<j} (\mathbbm{1}[ (\sigma_1^i > \sigma_1^j) \wedge  (\sigma_2^i > \sigma_2^j) ] + \mathbbm{1}[ (\sigma_1^i < \sigma_1^j) \wedge  (\sigma_2^i < \sigma_2^j)]). 
\end{equation}
$\tau( \sigma_1 ,  \sigma_2)$ counts the pairwise agreements between items from two rankings $\sigma_1$ and $\sigma_2$. $\overline{M} = \frac{1}{2} M(M-1)$ denotes total number of pairs. \mbox{$\tau$ ranges from $0$ (worst) to $1$~(best).}  

\paragraph{Baselines:} As for vanilla baselines, we first consider the vanilla Plackett-Luce model~\citep{plackett1975analysis,luce1959individual}. For the sake of fair comparison,  we optimize with two optimization approaches, i.e.,  gradient descent (PL)~\citep{boyd2004convex} and EM using data augmentation (PL-EM)~\citep{caron2012efficient}. 

In terms of robust RA, we compare the results with PeerGrader~\citep{raman2014methods}, which is a variation of the Plackett-Luce model for partial preferences while incorporating the user reliability estimation module. We also compare with the popular noisy ranking model CrowdBT~\citep{chen2013pairwise}. Since CrowdBT was originally designed for pairwise preferences, we generalize CrowdBT to partial preferences following rank-breaking~\citep{weng2011bayesian}. Namely, we first break each partial preference into a set of pairwise comparisons and then apply CrowdBT to each pairwise comparison independently. Note that we do not consider the robust RA proposed by \cite{han2018robust}, since it is tailor-design for the crowdsourcing setting. In particular, their method requires multiple preferences from each user for initializing the parameters, while it is not true for a more general setting considered in this work where only one preference from each user is usually available. 

In terms of our CoarsenRank, we consider all its three instantiations: CoarsenTH~(Equation~\eqref{CoarsenTH2}), CoarsenBT~(Equation~\eqref{CoarsenBT}) and CoarsenPL~(Equation~\eqref{CoarsenRank}). Similar to CrowdBT, we adopt the rank-breaking to generalize the CoarsenTH and CoarsenBT to partial preferences.

\paragraph{Calibration for real application} In real applications, the number of items involved in partial comparisons usually varies significantly. Some items (e.g., user-friendly cameras, or popular sushi) may appear frequently in the ranking list due to their popularity, while other items (e.g., professional cameras, or sushi with special taste) appear only in few preferences due to their professionality. In such cases, the final ranking will not be unique or even not converge. To ensure a unique solution and to avoid overfitting, regularization may be used. Specifically, we perform normalization over $\theta$, i.e., $\theta = C\cdot\theta/\sum_m \theta_m$. We fixed $C = M/2$ in our experiment for simplicity. The calibration method is applied to all EM-based approaches, i.e.,  CoarsenBT, CoarsenPL, and PL-EM.  In terms of other baselines, e.g., PL, CrowdBT and CoarsenTH, their formulation is 
a little different and the nonnegative constraint is no longer required. Therefore, the calibration method cannot be applied.  Following CrowdBT, we use virtual node regularization~\citep{chen2013pairwise}. Specifically, it augments the original dataset $\mathcal{R}_N$ with $\mathcal{R}_o$, which consists of the pairwise comparisons between all items and a virtual item $\theta_0$, namely $\mathcal{R}_o = \{\theta_m > \theta_0, \theta_m < \theta_0, m = 1, 2, \ldots, M\}$.

\subsection{Detailed descriptions of datasets}

We conducted our experiment on four real-world datasets introduced in previous research. The detailed descriptions of the datasets are introduced in the following.

The \textit{Readlevel} dataset~\citep{chen2013pairwise} contains English text excerpts whose reading difficulty level is annotated by workers from a crowdsourcing platform. This dataset consists of $490$ excerpts from $624$ workers, resulting in a total of $12,728$ pairwise comparisons. A total order for $490$ excerpts provided by the domain expert is regarded as ground truth. The number of annotations varies significantly for different workers, \mbox{ranging from $1$ to $42$.} 

The \textit{SUSHI} dataset is introduced in~\citep{kamishima2003nantonac}, which consists of partial preferences over $100$ types of sushi from
$n = 5,000$ customers. Following \citep{khetan2016data}, we generated the total order as ground truth using the vanilla PL over the entire $5,000$ preferences. To create training data, we randomly replaced $2,000$ preferences in the original \textit{SUSHI}  dataset with another $2,000$ random generated preferences. The number of preference provided by each customer is fixed to one. 

The \textit{BabyFace} dataset \citep{han2018robust} consists of the evaluations of workers from Amazon Mechanical Turk on images of children’s facial microexpressions from happy to angry,  which yields a collection of $3,074$ trinary preferences from $41$ workers. A total order over all microexpressions is provided as ground truth by the agreement of most workers after the experiment. \mbox{Each worker provides at least $60$~annotations.} 

The \textit{PeerGrading} dataset \citep{sajjadi2016peer} consists of assessments, i.e., Self grading and Peer grading, from $219$ students over $79$ group submissions. We then created the ordinal gradings by merging the Self grading and Peer grading regarding the same assignment provided by each student, which results in a total of $3,619$ preferences with each containing $2$ or $3$ items. Further, the TA gradings (following a linear order) provided by six teaching assistants over all submissions are considered as ground truth. The number of annotations from different students ranges from $2$ to $26$.

The statistics of four datasets are summarized in Table~\ref{statistics}.
\begin{table}[!htb]
	\centering
	\caption{\label{statistics} The statistics of four real ranking datasets}
	\renewcommand{\arraystretch}{1.2}
	\setlength{\tabcolsep}{1.3mm}{	
		\scalebox{0.97}{
	\begin{tabular}{|c|c|c|c|c|c|}
		\hline
		Dataset  & \#items ($M$) & \#users  & \#preferences ($N$)  & length of preferences ($k$)  & \#annotations per user\\\hline
		Readlevel   &   $490$ &  $624$   &   $12,728$     &     $2$     &     $1-42$        \\ 
		SUSHI  &   $100$ &  $5,000$ &   $5,000$       &     $10$      &      $1$          \\
		BabyFace       &   $18$  &  $41$    &   $3,074$      &     $3$     &    $\ge 60$     \\
	    PeerGrading  &   $219$ &  $79$    & $3,619$   & $2$ or $3$   &    $2-26$        \\ 
		\hline
	\end{tabular}}}
\end{table}

\subsection{Exploring the efficacy of the calibration method}

In section~\ref{experiment_setting}, we introduce a calibration method to the vanilla EM algorithm to deal with data sparsity. We claim that the calibration method is necessary when the collected preferences do not evenly cover the items.  In particular, the calibration method serves as regularization, which is helpful to avoid overfitting and leads to a unique solution.

\begin{figure}[!ht]
		\centerline{\includegraphics[width=0.67\textwidth]{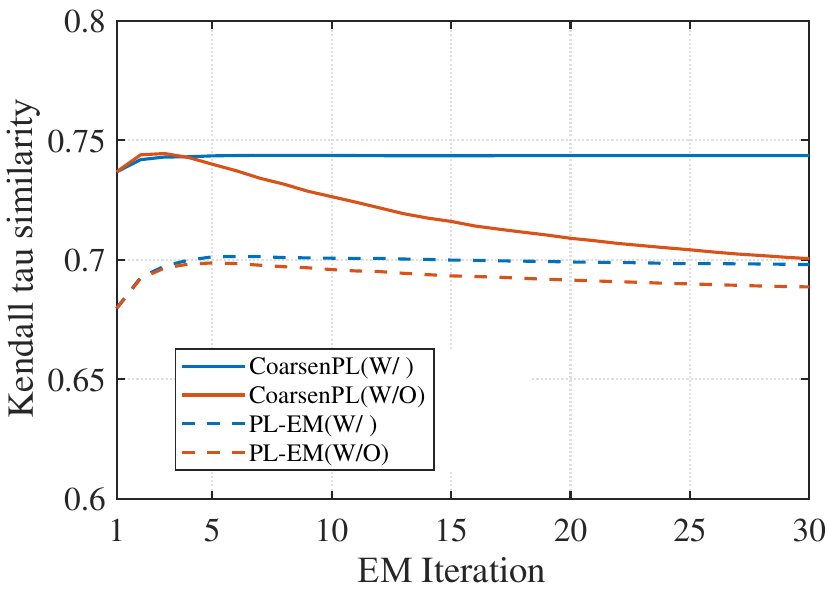}}
	\caption{\label{Cali} The performance comparison of CoarsenRank and PL-EM algorithm under the case of with or without calibration method. ``W/'' denotes ``with'' while ``W/O'' denotes ``without''.}	
\end{figure}

To verify the efficacy of the calibration method, we conducted rank aggregation using CoarsenPL and PL-EM on the \textit{Readlevel} dataset. Then we collected the Kendall tau similarity of these two methods in Figure~\ref{Cali}, under the case of with or without calibration method, respectively. 

Figure~\ref{Cali} shows that: (1) without the calibration method, CoarsenPL and PL-EM are all prone to overfitting. In particular, the Kendall tau similarity of CoarsenPL and PL-EM can reach their optima at the first few iterations but start to decrease in the later iterations. Since the Kendall tau similarity is different from our objective, this phenomenon is considered as a sign of overfitting.  (2) With the help of the calibration method, the overfitting problem is avoided. The Kendall tau similarity of two methods remains stable after reaching their optima, respectively. (3) With or without the calibration,  the optimum Kendall tau similarities of CoarsenPL are very close. The same is true for PL-EM. It implies that the calibration method does not change the result of rank aggregation methods but just avoids overfitting.

\subsection{ Deviance Information Criterion (DIC) for choosing the hyperparameter $\alpha$}

\begin{figure}[!ht]
	\begin{minipage}{0.49\linewidth}
		\centerline{\includegraphics[width=0.98\textwidth]{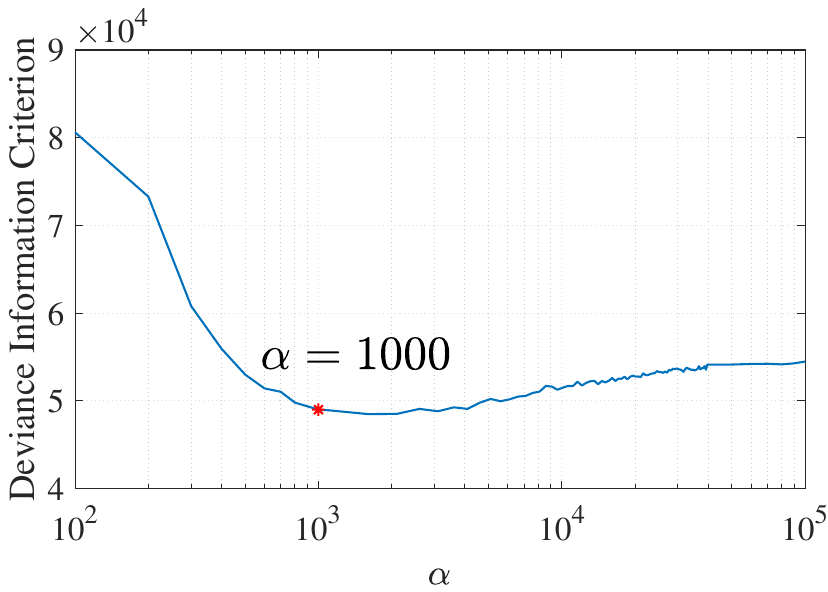}}
		\centerline{(a) Readlevel}
	\end{minipage}
	\hfill
	\begin{minipage}{0.49\linewidth}
		\centerline{\includegraphics[width=0.98\textwidth]{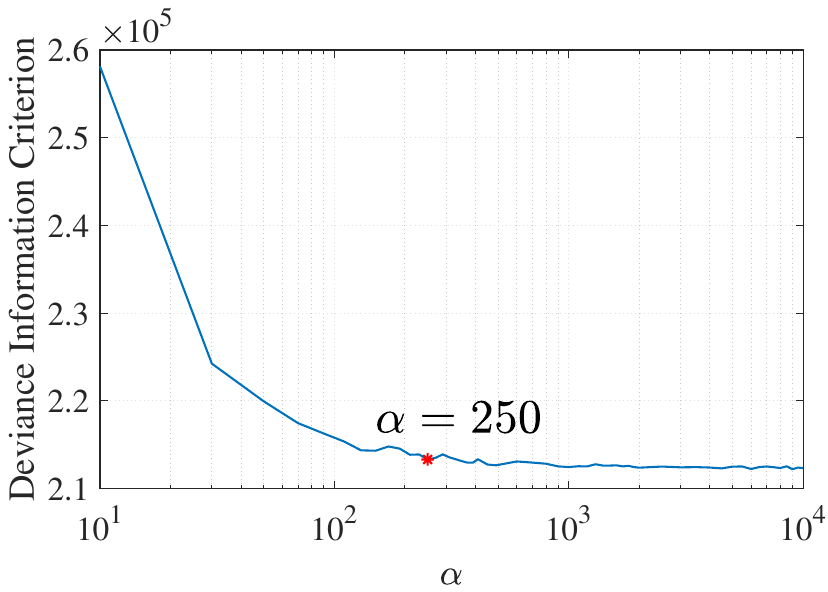}}
		\centerline{(b) SUSHI}
	\end{minipage}
	\\
	\begin{minipage}{0.49\linewidth}
		\centerline{\includegraphics[width=0.98\textwidth]{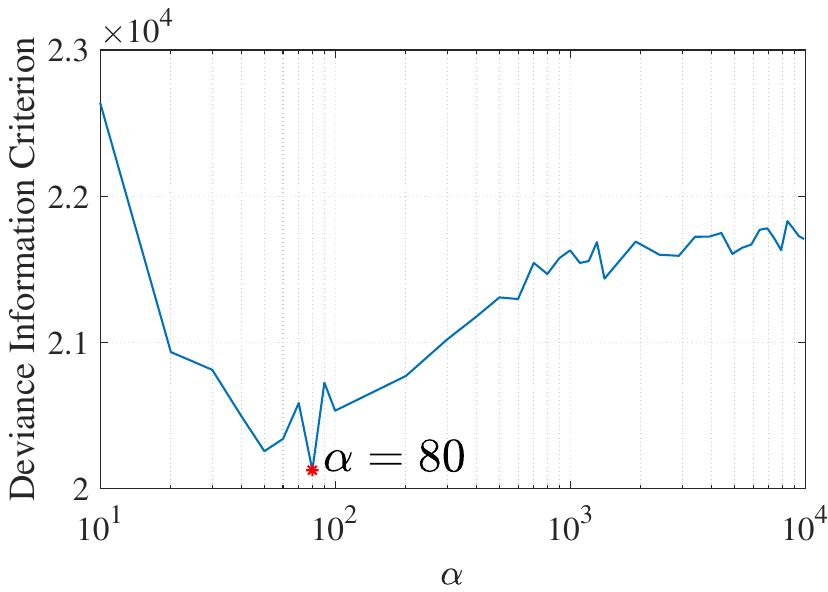}}
		\centerline{(c) BabyFace}
	\end{minipage}
	\hfill
	\begin{minipage}{0.49\linewidth}
		\centerline{\includegraphics[width=0.98\textwidth]{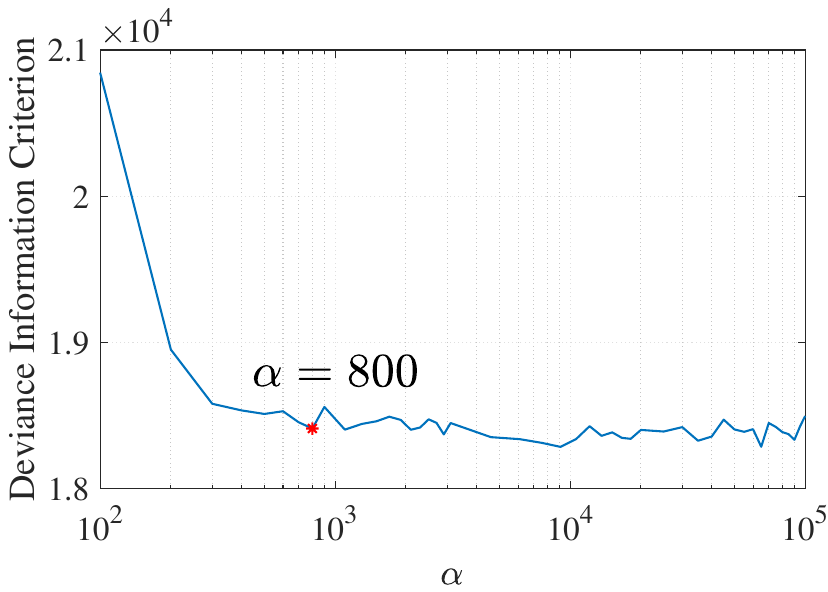}}
		\centerline{(d) PeerGrading}
	\end{minipage}		
   \\
	\caption{\label{Real_Data_ad}(a)-(d) The diagnostic plot of DIC VS. $\alpha$ on four datasets, respectively. The $\alpha$ used in our experiment are marked as ``{\color{red}{ $\ast$ }}'' in each figure.}	
\end{figure}

Following Section~\ref{DIC}, we adopted the DIC to choose the hyperparameter $\alpha$ for different datasets. Since it is intractable to analytically calculate the posterior expectation in DIC, we implemented a Gibbs Sampling procedure in Algorithm~\ref{CoarsenRank_Algo_Gibbs}. Then, we collected the samplings from~$P(\theta_m | \mathcal{R}_N, \Xi)$ (Equation~\eqref{theta}) and calculated the Monte Carlo estimation of DIC (Equation~\eqref{DIC_eq}) for different $\alpha$. The number of samplings is set to $50$ in our experiment. The diagnostic curves of $\alpha$ on four datasets are plotted in Figure~\ref{Real_Data_ad},~respectively. 

The results show that the DIC decreases dramatically at first when $\alpha$ is small, then the curve reaches a cusp and levels off, with more modest increases/decreases when $\alpha$ becomes larger. $\alpha$ is chosen at the point with the lowest DIC or where DIC levels off in our experiment, marked as ``{\color{red}{ $\ast$ }}'' in each figure.

\subsection{The accuracy of CoarsenRank in four real applications}
\begin{figure}[!ht]
		\centerline{\includegraphics[width=0.9\textwidth]{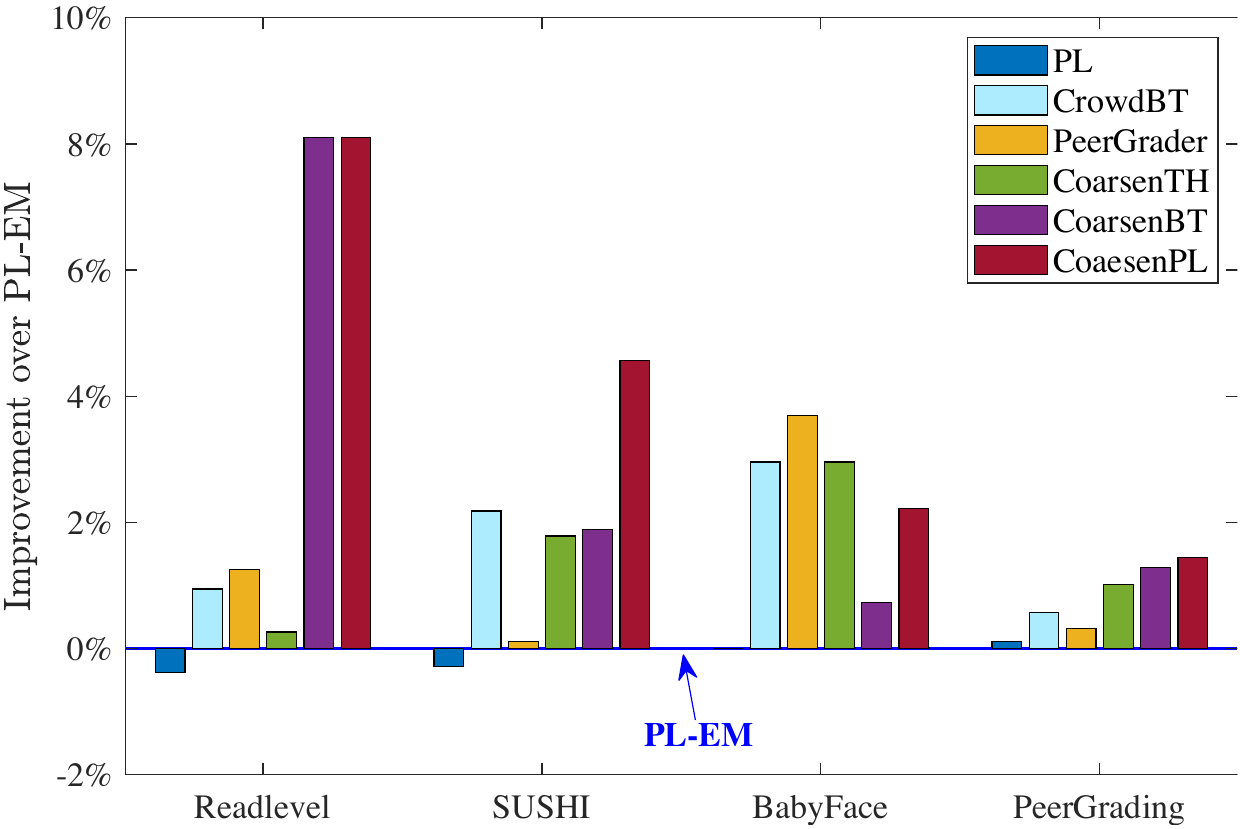}}
		\centerline{Performance}
	\caption{\label{Real_Data_e} Performance improvement of various methods over PL-EM on four datasets in terms of Kendall tau similarity (Eq.\eqref{acc_metric}), following $\frac{ \tau_ {\ast}  -  \tau_ 0} {\tau_ 0} $. $\tau_ 0$ is the Kendall tau similarity of PL-EM.}	
\end{figure}

We conducted the experiments on four real datasets and collected the empirical results of various rank aggregation methods in Table~\ref{Accuracy}. For better comparison, we further showed the performance improvement of all methods over PL-EM on four datasets in  Figure~\ref{Real_Data_e}.
\begin{table}[!t]
	\centering
	\caption{\label{Accuracy} Kendall tau similarity (Eq.\eqref{acc_metric}) of various rank aggregation methods on four real datasets. Best results are marked in bold. }
	\renewcommand{\arraystretch}{1.3}
	\setlength{\tabcolsep}{1.3mm}{	
		\scalebox{1}{
			\begin{tabular}{c|c|c|c|c|c|c}\toprule[1.3pt]
\multicolumn{2}{c|}{Baselines}         & Readlevel & SUSHI  & BabyFace & PeerGrading & Ave. Rank ($\downarrow$) \\\midrule[1pt]
\multirow{2}{*}{Vanilla} &PL         & 0.6853    & 0.8554 & 0.8824   & 0.8023 & 6.75     \\
                         & PL-EM      & 0.6879    & 0.8578 & 0.8824   & 0.8014 & 6.25     \\\hline
\multirow{2}{*}{Robust}  & CrowdBT    & 0.6944    & 0.8765 & 0.9085   & 0.8060 & 3     \\
                         & PeerGrader & 0.6965    & 0.8588 & \bf{0.9150} & 0.8040 & 3.5 \\\hline
\multirow{3}{*}{Coarsen} & CoarsenTH  & 0.6897    & 0.8731 & 0.9085   & 0.8095 & 3.5     \\
                         & CoarsenBT  & \bf{0.7436} & 0.8740  & 0.8889   & 0.8117 & 2.75     \\
                         & CoarsenPL  & \bf{0.7436} & \bf{0.8970} & 0.9020   & \bf{0.8130} & 1.75    \\\bottomrule[1.3pt]
\end{tabular}}}
\end{table}

\subsubsection{Comparison between three CoarsenRank variants and other baselines}

From Table~\ref{Accuracy} and Figure~\ref{Real_Data_e}, it can be observed that: (1)~Three CoarsenRank variants  achieve consistent improvement over other rank aggregation baselines. It demonstrates the great potential of CoarsenRank in real applications, where model misspecification widely exists. (2)~The accuracy of PL is comparable with PL-EM on all datasets, which rules out the possibility that the EM algorithm would lead to performance improvement. (3)~CrowdBT and PeerGrader get superior performance on \textit{BabyFace} because of sufficient annotations (over $60$) from each user and the trinary preferences setting in BabyFace. (4)~The improvement of CrowdBT and PeerGrader vary significantly on different datasets. The reason is that their pre-assumed perturbation patterns may not be consistent with noise agnostic perturbations in different datasets. (5)~Marginal improvement is achieved by CrowdBT and PeerGrader on \textit{Readlevel}, \textit{SUSHI} and \textit{PeerGrading} where each user provides almost one preference. Points $(4)$ \& $(5)$ are model misspecification cases which our CoarsenRank is intended to address.

\subsubsection{Comparison among three CoarsenRank variants}

From the fourth collum of Table~\ref{Accuracy}, we can find that: (1)~CoarsenPL achieves the highest Kendall tau similarity on most datasets, compared to other CoarsenRank variants. 
(2)~CoarsenBT gets the inferior performance to CoarsenPL. It is because CoarsenBT are tailored designed for pairwise preferences, and it needs to break each partial preference into independent pairwise preferences before aggregation, which has been recently shown to introduce inconsistency~\citep{khetan2016data}. (3)~CoarsenTH achieves the lowest Kendall tau similarity among all CoarsenRank variants. Apart from the above-mentioned inconsistency issue, CoarsenTH may suffer from accuracy reduction due to the introduced approximation or the lack of closed-form updating solutions.

\subsubsection{Ablation study on the number of annotations per user}
To justify our claim, we conducted an ablation study on the \textit{BabyFace} dataset by gradually reducing the number of annotations per user and collected the results in the following.

\begin{table}[!htbp]
\centering
\caption{\label{Ablation} Ablation study on the number of annotations per user using the \textit{BabyFace} dataset by gradually reducing the number of annotations per user. The Kendall tau similarity is calculated for various rank aggregation methods. Best results are marked in bold.}
\renewcommand{\arraystretch}{1.3}
\setlength{\tabcolsep}{1.3mm}{	
\scalebox{1}{
\begin{tabular}{c|c|c|c|c|c|c|c|c|c}\toprule[1.3pt]
\multicolumn{2}{c|}{Ratio}               & 1\%    & 2\%    & 4\%    & 8\%    & 16\%   & 32\%   & 64\%   & 100\%  \\
\multicolumn{2}{c|}{Average annotations} & 1.19   & 3      & 6      & 11.95  & 23.43  & 46.76  & 93.67  & 146.38 \\\midrule[1pt]
\multirow{2}{*}{Vanilla}  & PL  & 0.7778 & 0.7582 & 0.8497 & 0.8693 & 0.8824 & 0.8824 & 0.8824 & 0.8824 \\
& PL-EM  & 0.6928 & 0.6863 & 0.8235 & 0.8497 & 0.8497 & 0.8562 & 0.8824 & 0.8824 \\\hline
\multirow{2}{*}{Robust}   & CrowdBT& 0.7843 & 0.7908 & 0.8627 & \bf{0.8824} & 0.8824 & \bf{0.9085} & 0.9020 & 0.9085 \\
 & PeerGrader  & 0.8497 & 0.8562 & 0.8562 & \bf{0.8824} & \bf{0.9085} & \bf{0.9085} & \bf{0.9150} & \bf{0.9150} \\\hline
\multirow{3}{*}{Coarsen}  & CoarsenTH   & \bf{0.8824} & \bf{0.8824} & \bf{0.8824} & \bf{0.8824} & 0.8824 & 0.8824 & 0.8889 & 0.9085 \\
& CoarsenBT   & \bf{0.8824} & \bf{0.8824} & \bf{0.8824} & \bf{0.8824} & 0.8824 & 0.8824 & 0.8889 & 0.8889 \\
& CoarsenPL   & 0.8301 & 0.8431  & 0.8627 & 0.8693 & 0.8758 & \bf{0.9085} & 0.9085 & 0.9020 \\\bottomrule[1.3pt]
\end{tabular}}}
\end{table}

Table~\ref{Ablation} shows that: (1) the performance of three CoarsenRank variants is less sensitive to the number of annotations compared to other vanilla and robust rank aggregation methods; (2) when the number of annotation per user is insufficient (i.e., $\le 12$) to learn the ability of annotator, three CoarsenRank variants can achieve the best aggregation performance; (3) the performance of CoarsenPL is significantly better than its vanilla version, i.e., PL-EM, which demonstrates the superiority of CoarsenRank for noisy rank aggregation.

\subsection{Comparisons of all methods w.r.t. the learning time }

\begin{figure}[!t]
	\centerline{\includegraphics[width=0.9\textwidth]{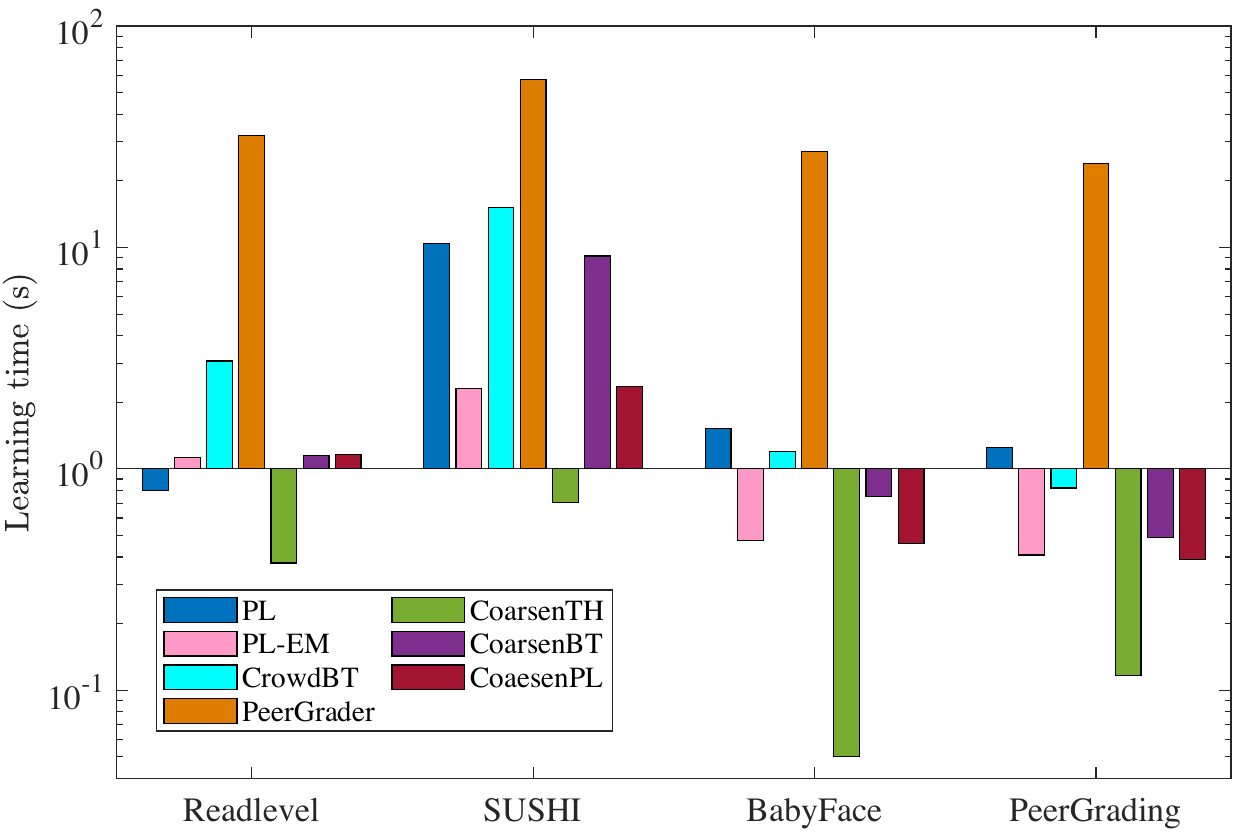}}
		\centerline{learning time (s)}
	\caption{\label{Real_Data_f}The learning time of all baselines and the proposed ones on four datasets, respectively.}	
\end{figure}

We independently ran each baseline $50$ times and reported the learning time (s) in Table~\ref{Computation} and Figure~\ref{Real_Data_f}. The learning time is represented by the mean with the standard deviation. For the sake of fair comparison, the inner iteration is fixed to $15$ for all methods. Empirical analyses were performed on an Intel i5 processor($2.30$ GHz) and $8$ GB random-access memory (RAM).
\begin{table}[!ht]
	\centering
	\caption{\label{Computation} The learning time (s) of all baselines and the proposed ones on four datasets, respectively. Best results are marked in bold.}
	\renewcommand{\arraystretch}{1.3}
	\setlength{\tabcolsep}{1.3mm}{	
		\scalebox{0.93}{
			\begin{tabular}{c|c|c|c|c|c|c}\toprule[1.3pt]
\multicolumn{2}{c|}{Baselines}   & Readlevel  & SUSHI  & BabyFace   & PeerGrading  & Ave. Rank($\downarrow$) \\\midrule[1pt]
\multirow{2}{*}{Vanilla} &PL         & $0.793 \pm 0.050$ & $10.416 \pm 1.605$   & $1.522 \pm 0.087$  & $1.250 \pm 0.075$  & 4.75 \\
& PL-EM      & $1.352 \pm 0.124$   & $2.305 \pm 0.338$  & $0.475 \pm 0.038$ & $0.408 \pm 0.037$ & 2.5 \\\hline
\multirow{2}{*}{Robust}  & CrowdBT    & $3.831 \pm 0.232$   & $15.169 \pm 1.598$   & $1.193 \pm 0.0554$  & $0.819 \pm 0.072$  & 5.5\\
& PeerGrader & $36.580\pm0.967$  & $57.289 \pm 1.347$   & $27.217 \pm 0.973$  & $23.966 \pm 0.128$ & 7 \\\hline
\multirow{3}{*}{Coarsen} & CoarsenTH  & $\bf{0.253 \pm 0.014}$ & $\bf{0.706 \pm 0.029}$ & $\bf{0.050 \pm 0.011}$ & $\bf{0.117 \pm 0.018}$ & 1 \\
& CoarsenBT  & $1.355 \pm 0.102$ & $9.155 \pm 0.308$  & $0.748 \pm 0.059$ & $0.488 \pm 0.016$ & 4\\
& CoarsenPL  & $1.364 \pm 0.098$ & $2.350 \pm 0.671$   & $0.459 \pm 0.012$ & $0.389 \pm 0.024$ & 3.25\\\toprule[1.3pt]
\end{tabular}}}
\end{table}


Table~\ref{Computation} and Figure~\ref{Real_Data_f} shows that: (1)~Three CoarsenRank variants achieve much smaller learning time compared to other robust ranking aggregation baselines. It shows that our CoarsenRank is promising for deploying in a large-scale environment, where reliability and efficiency are both required. (2)~The learning time of CoarsenPL and PL-EM are comparable because of the only difference between CoarsenPL and PL-EM lying at the choosing of parameter $\tau$ (See Equation~\eqref{theta_update}).  (3)~PeerGrader suffers from significant inefficiencies since it needs to optimize parameters alternatively. (4)~CrowdBT replaces the inefficient alternative optimization with the online Bayesian moment matching and achieves smaller learning time compared to PeerGrader.  However, it is still time-consuming on \textit{SUSHI} dataset because of the inefficient rank-break method for long preferences.

\begin{figure}[!t]
		\centerline{\includegraphics[width=0.85\textwidth]{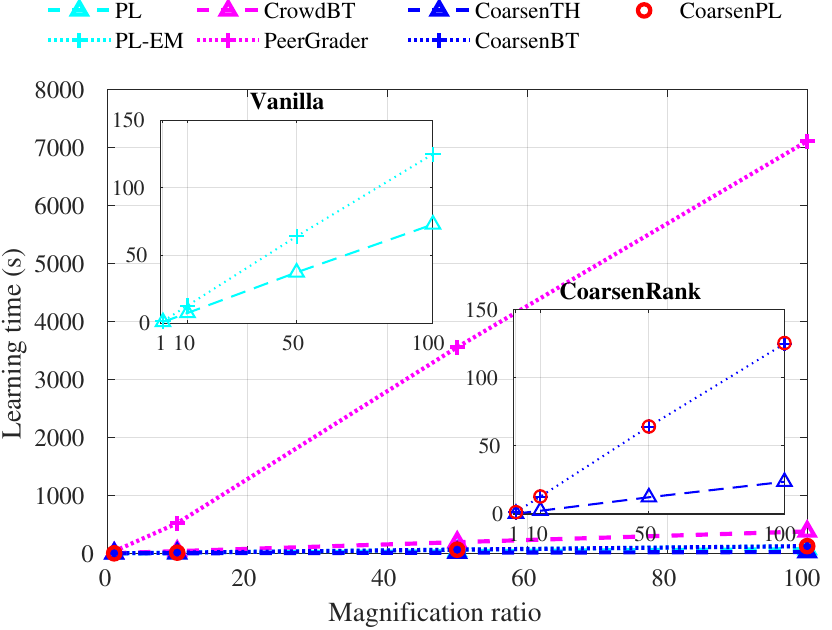}}
	\caption{\label{time_ratio} Visualization of the learning time (s) for various rank aggregation methods on the magnified \textit{Readlevel} dataset.}	
\end{figure}

\subsubsection{Ablation study on the size of samples}
To demonstrate the time complexity analysis in Section~\ref{time-space}, we took the \textit{Readlevel} dataset as an example for the experiment since it has the largest number of samples ($13k$). We first magnified the \textit{Readlevel} dataset by duplicating it with the specified ratios and then collected the learning time (s) of various methods on each magnified dataset, respectively. For the sake of fair comparison, the inner iteration is fixed to $15$ for all methods. The learning times are summarized in Table~\ref{Ablation_size} and Figure~\ref{time_ratio}, 

\begin{table}[!ht]
\centering
\caption{\label{Ablation_size} Ablation study on the size of the data using the \textit{Readlevel} dataset by gradually magnifying the size of data. The learning time (s) is represented by the mean with the standard deviation. Best results are marked in bold.}
\renewcommand{\arraystretch}{1.3}
\setlength{\tabcolsep}{1.3mm}{	
\scalebox{1}{
\begin{tabular}{c|c|c|c|c|c}\toprule[1.3pt]
\multicolumn{2}{c|}
{Magnification ratio} & 1   & 10    & 50    & 100   \\\midrule[1pt]
\multirow{2}{*}{Vanilla}  
& PL           & 0.793 $\pm$ 0.050  & 7.445$\pm$0.113 &  37.445$\pm$ 0.298 & 72.915$\pm$0.960\\
& PL-EM       & 1.352 $\pm$ 0.124 & 12.909 $\pm$ 0.286 & 64.081 $\pm$ 0.690  & 124.838 $\pm$ 2.222 \\\hline
\multirow{2}{*}{Robust}   
& CrowdBT     & 3.831 $\pm$ 0.232 & 38.703 $\pm$ 0.573 & 193.832 $\pm$ 1.105 & 378.552 $\pm$ 8.037 \\
& PeerGrader  & 36.580$\pm$0.967& 520.240$\pm$9.908 & 3557.707$\pm$21.783  & 7113.333$\pm$69.421\\\hline
\multirow{3}{*}{Coarsen}  
& CoarsenTH   & \bf{0.253 $\pm$ 0.014} &  \bf{2.367$\pm$0.028} & \bf{12.258$\pm$0.213} &\bf{23.650$\pm$0.383} \\
& CoarsenBT   & 1.355 $\pm$ 0.102 & 12.793 $\pm$ 0.212 & 64.111 $\pm$ 0.663  & 124.851 $\pm$ 1.887 \\
& CoarsenPL   & 1.364 $\pm$ 0.098 & 12.858 $\pm$ 0.222 & 64.176 $\pm$ 0.691  & 125.343 $\pm$ 2.704 \\\bottomrule[1.3pt]
\end{tabular}}}
\end{table}

From Table~\ref{Ablation_size} and Figure~\ref{time_ratio},  we can find that: (1) the learning time of all the methods exhib obvious linear correlation with the size of the dataset. This is consistent with our analysis in Section~\ref{time-space} that the learning time is linearly correlated with the number of samples. (2) The superiority of our CoarsenRank variants over PeerGrader is significant. In particular, for a large-scale dataset with $100*13k$ samples, CoarsenTH can output the result in less than one minute, while PeerGarder needs almost 2 hours to get the result. (3) CoarsenRank (i.e., CoarsenBT and CoarsenPL) brings almost no extra computation cost over its vanilla counterpart (i.e., PL-EM) regardless of the number of samples. 

\section{Conclusion}\label{Conclusion}

Our CoarsenRank performs imprecise inference conditioning on a neighborhood of the ranking dataset, which opens a new door to the robust rank aggregation against model misspecification. Particularly, a computationally efficient formula for CoarsenRank is derived, which introduces only one extra hyperparameter to vanilla ranking models.  
Experiments on four real applications demonstrate imprecise inference on the neighborhood of the preferences, instead of the original dataset, can improve the model reliability. It shows that our CoarsenRank has great potential in real applications, e.g., social choice, information retrieval, recommender system, etc, where model misspecification widely exists. We consider the rank aggregation problem where only one ground truth consensus full ranking exists. In terms of other applications for mixture rank aggregation~\citep{zhao2016learning}, the homogeneity assumption on users is no longer valid and we will extend our distributionally robust CoarsenRank to the scenario of heterogeneity users. Another promising direction for future research is to explore other divergence metrics for other statistical properties of rank aggregation.

\appendix
\section{Acknowledgments}\label{Acknowledgments}
IWT is supported by ARC under grants DP180100106 and DP200101328. WC is supported by the national NSFC (No.61603338 and No.11426202) and the Zhejiang Provincial NSFC (LY21F030013). GN is supported by JST AIP Acceleration Research Grant Number JPMJCR20U3, Japan. MS is supported by the International Research Center for Neurointelligence (WPI-IRCN) at The University of Tokyo Institutes for Advanced Study.

\section{Proof for Theorem~\ref{fundation_theory}}
\paragraph{Theorem 1} Suppose $D(\mathcal{R}_N, \Re_N)$ is an almost surely (a.s.)-consistent estimator\footnote{In probability theory, an event happens almost surely if it happens with probability one.} of $D(P_o, P_\theta)$, namely $D(\mathcal{R}_N, \Re_N) \xlongrightarrow[N\rightarrow  +\infty]{\text{a.s.}} D(P_o, P_\theta)$, where $F_N(x|\mathcal{R}_N) \rightarrow P_o$ and $F_N(x|\Re_N) \rightarrow P_\theta$ when $N \rightarrow +\infty$. Assume $\mathbb{P}(D(P_o, P_\theta) = \epsilon) = 0$ and  $\mathbb{P}(D(P_o, P_\theta) < \epsilon) > 0$, then we have 
\begin{equation*}
\mathbb{P}(\theta | D(\mathcal{R}_N, \Re_N) < \epsilon) \ \xlongrightarrow[N\rightarrow  +\infty]{\text{a.s.}}\ \mathbb{P}(\theta | D(P_o, P_\theta) < \epsilon),
\end{equation*}
for any $\theta \in  \Theta$ such that $\int  |\theta| \mathbb{P}(d\theta) < \infty$.

\paragraph{Proof:}
Since $\mathbb{P}(D(P_o, P_\theta) = \epsilon) = 0$, we have\footnote{$\mathbbm{1}(x)$ denotes the indicator function, which returns one when $x$ is true and zero, otherwise.} $\mathbbm{1}(D(\mathcal{R}_N, \Re_N) < \epsilon) \xlongrightarrow[N\rightarrow  +\infty]{\text{a.s.}} \mathbbm{1}(D(P_o, P_\theta) < \epsilon)$. Then we have $|\mathbbm{1}(D(\mathcal{R}_N, \Re_N) < \epsilon)| \le 1$ hold $\forall N>0$ since $0 \le \mathbbm{1}(\cdot) \le 1$. According to the dominated convergence theorem  \citep{billingsley2013convergence}, we have $\mathbb{P}(D(\mathcal{R}_N, \Re_N) < \epsilon) \xlongrightarrow[N\rightarrow  +\infty]{\text{a.s.}} \mathbb{P}(D(P_o, P_\theta) < \epsilon)$.

Similarly, we have $\theta\mathbbm{1}(D(\mathcal{R}_N, \Re_N) < \epsilon)) \xlongrightarrow[N\rightarrow  +\infty]{\text{a.s.}} \theta\mathbbm{1}(D(P_o, P_\theta) < \epsilon))$. Since $0 \le \mathbbm{1}(\cdot) \le 1$, $|\theta\mathbbm{1}(D(\mathcal{R}_N, \Re_N) < \epsilon))| \le |\theta|$, $\forall N>0$. Therefore, we have  $\mathbb{P}(\theta\mathbbm{1}(D(\mathcal{R}_N, \Re_N) < \epsilon)) \xlongrightarrow[N\rightarrow  +\infty]{\text{a.s.}} \mathbb{P}(\theta\mathbbm{1}(D(P_o, P_\theta) < \epsilon))$, according  to the dominated convergence theorem and $\int  |\theta| \mathbb{P}(d\theta) < \infty$. Above all, we have 
\begin{equation*}
\begin{aligned}
\qquad \mathbb{P}(\theta | D(\mathcal{R}_N, \Re_N) < \epsilon) & =\frac{\mathbb{P}(\theta\mathbbm{1}(D(\mathcal{R}_N, \Re_N) < \epsilon))}{\mathbb{P}(D(\mathcal{R}_N, \Re_N) < \epsilon)} \\
& \xlongrightarrow[N\rightarrow  +\infty]{\text{a.s.}}
\frac{\mathbb{P}(\theta\mathbbm{1}(D(P_o, P_\theta) < \epsilon))}{\mathbb{P}(D(P_o, P_\theta) < \epsilon)} 
 =\mathbb{P}(\theta | D(P_o, P_\theta) < \epsilon).\qquad\qquad \hfill\rule{2mm}{2mm}
\end{aligned}
\end{equation*}

\bibliography{sample}   
\end{document}